\documentclass[1p,12pt,authoryear]{elsarticle}

\usepackage[T1]{fontenc}
\usepackage[utf8]{inputenc}
\usepackage{lmodern}
\usepackage{microtype}
\usepackage{seqsplit}
\usepackage{graphicx}
\usepackage{booktabs}
\usepackage{float}
\usepackage{array}
\usepackage{amsmath}
\usepackage{amssymb}
\usepackage{xcolor}
\usepackage[hyphens]{url}
\usepackage[hidelinks]{hyperref}
\usepackage{tikz}
\usetikzlibrary{positioning,arrows.meta}

\journal{Expert Systems with Applications}

\makeatletter
\long\def\@makefntext#1{%
  \raggedright
  \setlength{\leftskip}{1.4em}%
  \noindent\llap{\hb@xt@1.4em{\hss\@makefnmark}}#1}
\makeatother

\newcommand{\system}{Mingbird}
\newcommand{\lrab}{LRAB}



\begin{document}

\begin{frontmatter}

\title{Harness Leverage Across Model Size:
A Ladder of Five Local Agent Harnesses from 0.75B to 4.7B}

\author[ustb]{Hao Wang\corref{cor1}}
\author[honor]{Ting Huang\corref{cor1}}
\cortext[cor1]{Co-corresponding authors who contributed equally to this work.\newline
\textit{Email addresses:} \mbox{wanghao994916959@163.com} (Hao Wang),
\mbox{huangting6@honor.com} (Ting Huang).}
\affiliation[ustb]{organization={School of Mathematics and Physics, University of Science and Technology Beijing}, city={Beijing}, country={China}}
\affiliation[honor]{organization={Honor Device Co., Ltd.}, city={Beijing}, country={China}}

\begin{abstract}
A harness turns a model into a system that completes tasks: it builds the
prompt, exposes tools and decides when work is done. Harness comparisons hold
the model fixed. Model comparisons hold the harness fixed. We vary both: five
shipped harnesses, 18 artifact-producing tasks, four locally deployable
open-weight models from 0.75B to 4.7B and a 35.5B anchor, with one
deterministic checker scoring all 360 ladder cells. The completion-controlling
harness, \system{}, has the highest mean at every size. Of four pre-registered
paired contrasts, two survive Holm correction, at 0.75B and 4.7B. Thirteen of
sixteen paired intervals exclude zero under the deployment reading, with leads
from 0.075 over the strongest rival to 0.819 over the weakest. The completion
gate is not the carrier: disabling all six executable checks moves the mean by
$+0.080$, $+0.117$, $+0.003$ and $-0.004$, every interval including zero, and
only at 1.9B does disabling it create empty runs. Removing one repeat-blocking
rule alone costs 0.411 at 0.75B and leaves eight of eighteen tasks with no
artifact. The harness effect is a progress effect: it decides whether a run
leaves anything on disk. Of 357 usable cells, 127 leave none of their declared
artifacts and 85 of those exit normally, so a scorer that reads exit status
credits work that was never written. Whether the effect grows or shrinks with
size, four rungs cannot settle: a floor arm produces a rise, a ceiling a
decline, and a cross-family rung a peak.

\end{abstract}


\end{frontmatter}
\enlargethispage{2\baselineskip}

\bibliographystyle{apalike}
\section{Introduction}
\label{sec:intro}

A local agent is two things at once. One is a language model running on the
user's own hardware. The other is the software around it, which builds the
prompt, exposes tools, executes each call, feeds the result back, and decides
when the work is finished. In deployed systems this second layer is called the
harness, and its components are the ones a cognitive architecture for language
agents names: memory, action space and decision procedure \citep{sumers2024coala}.
For a frontier model reached over an API, the harness is largely a
convenience. For a model of between 0.75 and 4.7 billion parameters sharing an
8\,GB graphics card with its own context window, it can be what decides whether
the task is completed at all.

That regime is the one a single consumer card can serve. Small open-weight models
fit on such a card with a usable context window, and they are the models that
individuals and small teams run. The harnesses those users install were
written for a different regime. Two of the four third-party harnesses we test
are product-grade tools designed around strong models with reliable tool-call
syntax and long, self-correcting reasoning traces. One is a minimal construction
kit that trusts the model with four tools. One is a compact bash-only control
that assumes the model can hold a shell protocol. A small model breaks these
assumptions in different ways, and the resulting failures are not always
visible. An agent can end its turn with none of the deliverables written while
the process exits normally and the transcript contains no error.

\system{} was built against those assumptions being false. Its design point is
completion control: the harness does not accept the model's claim that the work
is finished until executable checks on the working directory agree. The
architecture and the full mechanism set are described in an earlier paper
\citep{mingbird2026}. This paper measures what that design point is worth, and
how the answer depends on the size of the model underneath.

Studies that compare harnesses hold the model fixed, and studies that compare
models hold the harness fixed. Neither design answers the question a
practitioner with one graphics card faces, which is whether to spend the card on
a larger model or on better software around a smaller one. The same trade-off
has a name in the scaling literature, where spending more inference compute can
substitute for parameters \citep{snell2024testtime}. A harness is the part of
that compute a user buys once and reuses. Recent studies do cross the two axes,
and they disagree. A pre-registered comparison of three
scaffolds on five models found the most capable model gaining the most. A
ten-model comparison reports diminishing harness gains as capability rises on
issue-repair work and the opposite on open-ended repository work. A
seven-scaffold study reports cost differences that widen on a local 27B model,
and a three-level comparison on 2--3B models finds that a minimal shell can be
worse than raw prompting. None of them holds five shipped harnesses, four locally
deployable sizes and one artifact-level scorer fixed at once.
Section~\ref{sec:related} sets these against each other.

We report a controlled comparison on one machine, one runtime, one task suite,
and one scorer. Five harnesses run 18 artifact-producing tasks against four
open-weight models from 0.75B to 4.7B, and a 35.5B mixture-of-experts model
anchors the top of the range on a six-task subset. Section~\ref{sec:design}
sets out the cells this runs to, and every cell is scored by a deterministic
checker that reads only files in the working directory.

\system{} has the highest mean at each of the four sizes. Four of the sixteen
paired comparisons were registered in advance and enter the Holm family, one
against goose at each rung: two survive correction, at 0.75B (adjusted
$p=0.0067$) and 4.7B ($0.0410$), and two do not. The remaining twelve are
descriptive, and thirteen of the sixteen exclude zero under the deployment
reading, with leads that run from 0.075 over the strongest rival at 4.7B to
0.819 over the weakest. The advantage is not the completion gate the harness was
built around: with all six executable checks switched off, the score moves by
$+0.080$ at 0.75B and $+0.117$ at 1.9B, both intervals including zero, and by
nothing at the two larger rungs. The one rung where the gate does its job is
1.9B, where switching it off leaves three of eighteen tasks with no artifact
instead of none. The harness changes whether a run ends with any
work product. Whether that effect grows or shrinks with model size is a
question this ladder cannot settle, and Section~\ref{sec:shape} shows the three
measurement artifacts that make an answer look available when it is not.

The decomposition says where the advantage lives. At 0.75B between half and
three quarters of each gap is cells the baseline
left empty, and most of the remainder is how far a completed run gets. At 1.9B
almost all of the gap is availability. Against pi at 4.7B, where no cell is
empty, the lead is a depth difference on tasks both arms finish. This design does
not decompose that machinery further, but it rules out the completion gate as
the carrier at three of the four rungs. The gate is the mechanism the system was
built around, and at the fourth rung it does what it was designed to do.

The machinery shows in how runs end. Of 357 usable cells,
127 leave none of their declared artifacts, and 85 of those 127 end with a
normal process exit. The rate runs from 6\% for \system{} to 69\% for
mini-swe-agent. A scoring rule that read the exit status as completion would
credit all 85 with work they never wrote, and a harness that reads it that way
inherits the model's stop as a task failure.

Three features of the design set the scope of these claims. An arm whose reply
contract no small model meets widens the five-arm span by construction, so the
span without it is reported beside the raw one. A scorer ceiling binds at the
top, where \system{} and goose both reach 1.000 on eight of eighteen tasks. And
the 3.7B rung is a different model family, so it is read as a deployment point
and not as a step on a capability scale. Section~\ref{sec:shape} reports what
these features do to the shape of the advantage. One further check runs a single
base model at four bit widths from 3.15 to 6.50 bits per weight, because the
ladder does not hold quantization constant. It locates where the advantage
survives compression and where it does not.

\section{Related work}
\label{sec:related}
Three bodies of work bear on this study: how agent loops are built and measured,
what is already known about harness and scaffold effects, and the resource
envelope of local inference.

\paragraph{Agent loops and their benchmarks} The dominant recipe descends from
ReAct \citep{react}, in which the model interleaves reasoning with tool calls
and the harness executes each call and appends the observation. Practitioner
syntheses describe the same loop as a small set of composable patterns
\citep{anthropic_agents}. Toolformer \citep{toolformer}, Gorilla \citep{gorilla}
and ToolLLM \citep{toolllm} established learned tool use and reliable API
invocation. Reflexion \citep{reflexion} and Self-Refine \citep{selfrefine}
improve behavior inside the loop through self-critique, and both presuppose a
model that can produce a usable critique of its own attempt. Self-correction
without external feedback often fails \citep{huang2023selfcorrect,
kamoi2024selfcorrect}, while CRITIC \citep{critic} shows that supplying tool
feedback repairs the loop. That is the design choice behind the completion
guards measured here: they execute checks instead of asking the model to
introspect \citep{mingbird2026}.

Benchmarks for agents span environments and interaction styles. \mbox{AgentBench}
\citep{agentbench}, AgentBoard \citep{agentboard}, WebArena \citep{webarena},
OSWorld \citep{xie2024osworld}, SWE-bench \citep{swebench} and GAIA \citep{gaia}
cover browsing, desktops, repositories and open-ended assistance. $\tau$-bench
\citep{taubench} and $\tau^2$-bench \citep{tau2bench} add a simulated user and
verify final database state. AndroidWorld \citep{androidworld} and AppWorld
\citep{appworld} move to mobile and application environments. ToolSandbox
\citep{toolsandbox} and CRMArena \citep{crmarena} test stateful tool use, and
AgentDojo \citep{agentdojo} and CaMeL \citep{camel} add prompt-injection safety.
These suites vary the model or the task. The suite used here, \lrab{}, was
introduced in an earlier paper \citep{mingbird2026} and varies the harness on
artifact-producing tasks.

\paragraph{Harness and scaffold effects} That the software around a model changes
measured performance is established, but its dependence on model size is
contested. One line of work names the binding-constraint thesis and calls for
mandatory harness disclosure \citep{harnessdisclosure}. A pre-registered
three-scaffold, five-model comparison on GAIA finds within-model scaffold effects
up to 28 points, and its prediction that better models would be less
scaffold-sensitive was rejected in direction: at the harder level the most
capable model gained the most \citep{scaffoldgaia}. A ten-model comparison of
mini-swe-agent and OpenCode reports diminishing harness gains as capability
rises on issue-repair benchmarks, with the opposite pattern on open-ended
repository work \citep{hu2026beyond}. On the cost side, seven scaffoldings
against five models on one fixed software task show that scaffolding dominates
cost, with the widest spread on a local 27B model \citep{alier2026scaffold}. A
theory result points the other way: an inference-time harness can compute a
closure its per-call primitive cannot, and the gap closes by enlarging the model,
easing the task, or lengthening the chain of thought \citep{gai2026separations}.
That predicts a harness effect that shrinks with size on tasks a larger model can
represent directly. One very recent system study reads the balance the other way
again: across its traces, task accuracy depends mostly on the model, while
harness-side variance moves cost by up to 2.6 times \citep{park2026cost}. Its
measurements come from Claude Code and Codex, harnesses built around frontier
models, on software benchmarks. The band below 5B, where a run can end with no
file on disk, is the band this paper measures.

Five studies come closest to this one. A comparison of three harness levels on
three 2--3B models over 24 tasks reports a non-monotonic result in which a
minimal shell can be worse than raw prompting \citep{cho2026size}. The ten-model
study above covers the capability range and gives the closest evidence on
direction. A survey of 66 model-harness configurations, including pi, on frontier
models finds model rankings reversing across harnesses \citep{fit2026}. A
full-factorial comparison of six open-source harnesses and three lightweight
coding models over 100 SWE-bench Verified tasks reports resolution rates that
differ by up to 44 points across harnesses for one model, rankings that change
across models, and more than a 20-fold spread in tokens per resolved task
\citep{pairingmatters}. The CAIS Arena pins one model across six frameworks on a
shared tool server \citep{caisarena}. Our design differs in the band it covers,
0.75B to 4.7B on one consumer card, in scoring declared artifacts rather than
test patches, and in a completion-control ablation repeated at four deployment
points, which turns a curve into a mechanism.

Two related lines supply mechanisms, not comparisons. Loop detection goes
back to the StuckDetector in OpenHands \citep{openhands}, which normalizes events
and counts repeated action-observation and action-error patterns, and the
duplicate-call intercept measured in Section~\ref{sec:failure-results} follows
that paradigm. Agentless \citep{agentless} shows that a deliberately narrow
scaffold can outperform agentic ones on software-engineering tasks, and SWE-agent
\citep{sweagent} makes the agent-computer interface a design object in its own
right. ADAS \citep{adas}, AFlow \citep{aflow}, GPTSwarm \citep{gptswarm} and DSPy
\citep{dspy} search harness structure automatically, and AI Agents That Matter
\citep{agentsthatmatter} argues for cost-controlled evaluation. This study varies
whole harnesses, not interfaces, and fixes budgets across arms.

\paragraph{Local inference and small models} llama.cpp \citep{llamacpp}
established efficient quantized inference on commodity hardware, and Ollama
\citep{ollama} packages it behind a serving API. Surveys of small language models
\citep{slmsurvey} document the latency and memory envelope, and one position
paper argues that small models are the future of agentic AI
\citep{belcak2025slmagents}. SmolLM2 \citep{smollm2} and the Gemma and Qwen
families
\citep{gemma,qwen3tr,qwen25} push capability per parameter. Long-context
degradation is documented by Lost in the Middle \citep{lostmiddle}, Context Rot
\citep{contextrot} and same-task-more-token analyses \citep{sametaskmoretokens}.
Constrained decoding \citep{outlines,xgrammar,jsonschemabench} targets the
tool-call format failures that dominate small-model runs, and format-restriction
studies \citep{tam2024speak} show the trade-off. Lightweight agent libraries
\citep{smolagents,openinterpreter} make small models drivable with minimal
scaffolding, and TinyAgent \citep{tinyagent} explores function calling at the
edge. The two product harnesses tested here \citep{goose,opencode} are designed
around large-context cloud models and Unix shells, and the two minimal ones
\citep{piclient,miniswe} assume a model that can drive a shell unaided.
\system{}
starts from the opposite end: 1--9B models, a Windows host, a consumer GPU, and
tasks that mix code, files and research. In most empirical work on agent
frameworks the harness is a constant. This study makes it the variable.

\section{Harnesses, models, and protocol}
\label{sec:design}
The study compares five harnesses on one task suite, one scorer, one runtime and
one host, across four model sizes. The design is a factorial matrix over models
and harnesses, plus a mechanism gradient that repeats one completion-control
ablation at four sizes, plus a single upper anchor. It runs to 360 ladder cells,
a 216-cell gradient at the same four rungs with a 54-cell repeat of its 3.7B
rung, a 72-cell repeat of the ladder's two decisive rungs, a 36-cell loop-guard
ablation at 0.75B, 36 anchor cells of which the twelve on the six-task subset are
reported, and a 144-cell precision ladder.

\subsection{The completion-controlling harness}
\label{sec:mingbird}

\system{} is a harness written for models in the 1B to 9B range on a Windows
host with a consumer GPU. Its architecture, its seven design principles, and its
full mechanism set are described in an earlier paper \citep{mingbird2026}, and
this section states only what the measurements need. A \system{} run is a
plan-execute loop with a typed tool surface, and its distinguishing feature is a
completion gate. Before the harness accepts the model's claim that the work is
finished, six executable checks run at completion time: a no-work refusal, a
test guard, a finish-claim artifact check, a plan-completeness gate, an
artifact-content check, and a post-finish summary request. All six are disabled
by \texttt{AGENT\_ABLATION=finish\_gate}. A seventh, non-executable intervention,
a todo-sync reminder, stays on in the full and guards-off arms and is disabled in
the text-only arm. A textual re-read of the task stays on in every arm. The gate
is the design hypothesis under test: if an agent's characteristic failure is a
normal exit with the work unfinished, then a harness that checks the work instead
of trusting the claim should recover part of the loss.

\subsection{Models}

Four open-weight models span 0.75B to 4.7B, and all four fit entirely in the
8\,GB card at the study context. Three come from Qwen3.5 \citep{qwen35} at 0.75B,
1.9B and 4.7B, which gives a within-family ladder that holds tokenizer and chat
template constant. Granite 4.2 at 3.7B \citep{granite42} is a cross-family
anchor. Table~\ref{tab:bmodels} lists the parameter counts, architectures, and
served quantizations, all read from the served artifacts and not from model
names. A 35.5B mixture-of-experts model \citep{ornith15} with about three billion
active parameters anchors the top of the range on a reduced task set. It is an
anchor in parameter count and deployment cost, not in measured capability: in
the earlier matrix the 4.7B Qwen outscored it over the same eighteen tasks
\citep{mingbird2026}.

Parameter count does not order capability across families, so a cross-family rung
can be weaker than the rung below it. We therefore read the ladder as a set of
deployment points, not a capability scale, and the analysis reports
the within-family slope alongside the four-rung one for exactly this reason.

\newcommand{\tabbodyC}{%
\small
\begin{tabular}{@{}lllllll@{}}
\toprule
Role & Tag as served & Parameters & Architecture & Quantization & File & Family \\
\midrule
0.75B rung & \texttt{qwen3.5:0.8b} & 0.75B & dense & Q8\_0 & 1.3\,GB & Qwen3.5 \\
1.9B rung & \texttt{qwen3.5:2b} & 1.9B & dense & Q8\_0 & 2.7\,GB & Qwen3.5 \\
3.7B anchor & \texttt{granite4.2:3b} & 3.7B & dense & Q4\_K\_M & 2.2\,GB & Granite 4.2 \\
4.7B rung & \texttt{qwen3.5:4b} & 4.7B & dense & Q4\_K\_M & 3.4\,GB & Qwen3.5 \\
35.5B anchor & \texttt{ornith-1.5:35b} & 35.5B (3B active) & MoE & Q4\_K\_M & 22.6\,GB & Qwen3.5-MoE \\
\bottomrule
\end{tabular}}
\newlength{\fitwC}
\settowidth{\fitwC}{\tabbodyC}
\begin{table}[t]
\centering
\caption{The model ladder. Parameter counts, quantization, and file sizes are
read from the served artifacts (\texttt{ollama show}, \texttt{ollama list})
rather than from model names. File sizes are the served artifacts' byte counts
in decimal gigabytes, which is the basis the server's own listing uses. A binary
unit tool reports the same files about 7 percent smaller. Architecture is the
served GGUF architecture, and
the two architectures differ: qwen35 and granite are dense decoders, and the
35.5B model is a mixture of experts that routes each token to eight of its 256
experts (the served artifact's \texttt{expert\_used\_count}), which is where its
active-parameter figure comes from. All models run on one Ollama 0.40.6 server
at a 32K context. The Qwen3.5 rows form a within-family ladder and the 3.7B
Granite row is a cross-family anchor. The 35.5B anchor is analyzed separately. Quantization
follows each published tag and is therefore not held constant: the two smallest
Qwen tags ship at Q8\_0, the others at Q4\_K\_M, and Section~\ref{sec:precision}
measures what that change is worth. Model
sources: \citep{qwen35,granite42,ornith15}.}
\label{tab:bmodels}
\resizebox{\linewidth}{!}{\begin{minipage}{\fitwC}\tabbodyC\end{minipage}}
\end{table}

\subsection{Harnesses}

Five harnesses span the design space. \system{} is the completion-controlling
harness of Section~\ref{sec:mingbird}. pi is a minimal, MIT-licensed
construction kit with a four-tool loop. OpenCode and goose are product-grade
harnesses with native JSON tool calling and different architectures, a
client-server product and a protocol broker. mini-swe-agent is the compact
bash-only control used in recent harness studies. Table~\ref{tab:bharnesses}
lists versions and the tool-call path of each. Table~\ref{tab:bconfig} lists how
sampling, thinking, context and the wall budget are enforced for each arm,
because those mechanisms differ by harness and are part of the treatment.

\newcommand{\tabbodyB}{%
\small
\setlength{\tabcolsep}{3pt}
\begin{tabular}{@{}ll>{\raggedright\arraybackslash}p{2.3cm}>{\raggedright\arraybackslash}p{2.3cm}>{\raggedright\arraybackslash}p{2.6cm}@{}}
\toprule
Harness & Version & Tool-call path & Role in the design & Stop rule \\
\midrule
\system{} & v3.1.9 (56cd23d) & native JSON plus shell tools & small-model-tuned & model-decided plus executable checks \\
pi & 1.3.0 & native JSON (bash, read, write, edit) & minimal construction kit & model-decided, no checks \\
OpenCode & 1.18.23 & native JSON & product harness & model-decided, no checks \\
goose & 1.48.0 & native JSON, MCP/ACP broker & product harness & model-decided, no checks \\
mini-swe-agent & 2.4.6 & bash actions & compact bash-only control & model-decided, step limit, no checks \\
\bottomrule
\end{tabular}}
\begin{table}[t]
\centering
\caption{The harness set. ``Tool-call path'' is how the model's action reaches
the environment. ``Stop rule'' is how a run ends: the baselines stop when the
model says it is finished (with a step or turn limit in the runner), while
\system{} also runs executable checks before accepting a finish claim. The two
paths are a design axis, not a quality ranking.}
\label{tab:bharnesses}
\tabbodyB
\end{table}

\newcommand{\tabbodyA}{%
\footnotesize
\begin{tabular}{@{}ll>{\raggedright\arraybackslash}p{3.4cm}>{\raggedright\arraybackslash}p{3.4cm}>{\raggedright\arraybackslash}p{3.0cm}@{}}
\toprule
Arm & API path & Temperature & Thinking off & Context \\
\midrule
\system{} & Ollama \texttt{/api/chat}, direct & \texttt{AGENT\_TEMP} = 0 & \texttt{AGENT\_THINK} = 0 & \texttt{AGENT\_CTX} = 32768, compaction 8000 \\
pi & OpenAI \texttt{/v1} via proxy & proxy rewrites to 0 & \texttt{-{}-thinking off} & server default 32768 \\
OpenCode & OpenAI \texttt{/v1} via proxy & proxy rewrites to 0 & proxy sets \texttt{reasoning\_effort} = none & server default 32768 \\
goose & Ollama provider via proxy & \texttt{GOOSE\_TEMPERATURE} = 0, proxy rewrites & proxy sets \texttt{reasoning\_effort} = none & server default 32768 \\
mini-swe-agent & OpenAI \texttt{/v1} via proxy & \texttt{temperature} = 0 in the model call & proxy sets \texttt{reasoning\_effort} = none & server default 32768 \\
\bottomrule
\end{tabular}}
\newlength{\fitwA}
\settowidth{\fitwA}{\tabbodyA}
\begin{table}[t]
\centering
\caption{How each arm is configured. The protocol pins sampling, thinking, and
context at the transport, because the harnesses expose different knobs. The
API-path and Context columns say what enforces each pin for that arm. All arms
run against the same Ollama 0.40.6 server, and all of them end a cell by killing
its process tree at the wall budget, while mini-swe-agent's runner additionally allows
120\,s per command.}
\label{tab:bconfig}
\resizebox{\linewidth}{!}{\begin{minipage}{\fitwA}\tabbodyA\end{minipage}}
\end{table}

The unit of comparison is the harness as shipped, not a prompt written for this
study. Each harness brings its own system prompt, tool surface, context policy
and completion logic, and that bundle is the artifact a practitioner installs.
Holding prompts constant across products would measure a configuration none of
them ships and would remove the mechanisms under study. The price is that a
difference between two harnesses cannot be attributed to one design decision. The
mechanism gradient exists to pay that price, because it switches the
completion-control family on and off inside one harness with everything else
fixed.

The five harnesses also differ in tool-call protocol. Four pass a structured call
from the model to a typed tool. mini-swe-agent drives the shell with text and
runs whatever bash block the model writes. The protocol axis is not separable
from the arm's role here, because the only text-driven harness is also the only
bash-only control, so we treat the axis as descriptive, not as a factor.

\subsection{Tasks and scoring}

The task suite is an 18-task held-out set: fifteen workflow tasks (\mbox{HWF-01--15})
and three long-horizon tasks (HLH-01--03). It was frozen before the ladder
started and covers four tiers: retrieval over a fixture database (\mbox{HWF-01--02}),
synthesis across records (HWF-03--06), multi-step workflows with file edits and
checks (HWF-07--15), and long-horizon runs with an intermediate deliverable
(HLH-01--03). Each task ships a fixture directory, a prompt, and a deterministic
scorer. The scorer reads only files in the working directory and awards partial
credit per check. It never reads the agent transcript, so a harness cannot score
by describing work it did not do. The scorer's optional judge dimension is
disabled in every reported run. The score is milestones at 4/7 and final
artifacts at 3/7. Table~\ref{tab:btasks} in the appendix lists the tasks and the
number of checks each carries.

The held-out set mirrors a development set task for task: HWF-01 is a
literature-map task where WF-01 is a literature review, HWF-05 a cleaning
pipeline where WF-05 is one, and so on. The instances are distinct, with
different fixtures, prompts, required artifacts and checks. The families are not
distinct, and the design is deliberate: a family-level match lets the ladder ask
whether the harness effect depends on model size and not on one particular
fixture, while the development cells stay untouched. Every cell reported here is
a new run under this protocol.

\subsection{Execution protocol}

Every arm runs on the same Ollama 0.40.6 server with temperature zero and
thinking disabled. Context is pinned at 32K for every arm: \system{} through
\texttt{AGENT\_CTX}, with its compaction summarization call at 8000 as
Table~\ref{tab:bconfig} records, and the four CLI arms, which send no context
field, through the server's \texttt{OLLAMA\_CONTEXT\_LENGTH} default of 32768
fixed at the server restart of 22 October 2026. The loaded model's processor
split and window are recorded from \texttt{/api/ps} at the end of every cell, so
a cell that ran partly on the CPU, or against a server whose window was not the
protocol's, can be identified and not merely inferred. The record reaches every
scored cell except the first three, whose window rests on the server log rather
than on a per-cell snapshot.

The wall budget is set per (task, model) and is identical across harnesses for a
given task and model. Each budget is $1.3\times$ the maximum wall clock observed
for that cell among the completed cells of an earlier collection on the same
machine, floored at 10 minutes and capped at 120. Where fewer than three cells
of that task and model completed, the per-tier value applies instead, 20 minutes
at 0.75B, 30 at 1.9B and 3.7B, and 45 at 4.7B, plus 30 for the long-horizon
tasks. A single budget per tier would be wrong in both directions, because the
spread across tasks exceeds the spread across models: on granite4.2:3b the
maximum completed wall clock is 3.9 minutes on HWF-12 and 50.5 on HLH-03. Within a rung the order of harnesses alternates by task,
forward for odd tasks and reverse for even ones, so that warm-up or drift cannot
systematically favor one harness.

The 35.5B anchor ran all eighteen tasks at a 45-minute budget per cell and is
reported on the six with the fewest final deliverables, ties broken by task id. The
budget is set above this model's observed needs rather than to a round number,
and one cell at this rung still reaches the 45-minute limit. Only a
quarter of the model's weights are resident in the card (5.5 of 22.6\,GB), so
the rung runs mostly on the CPU and its wall clock is not comparable to the
ladder's.

\subsection{What the instrument can measure}

We report four measurement properties of the suite, all from the collected
cells and with no additional runs.

\emph{Discrimination.} A task separates harnesses at a rung if the five arms do
not all land on the same score. At 0.75B, 16 of 18 tasks separate, and two
(HLH-02 and HWF-05) return zero for every arm. At 1.9B every task
separates, and at 3.7B 17 of 18 do, HWF-01 being zero for every arm. Dead tasks
cost nothing in a paired contrast, because they are zeros for every arm, but the
nominal 18-task coverage is 16 tasks at the smallest rung.

\emph{Reliability.} We split the 18 tasks at random into two halves, recompute
the five-arm profile on each half, and correlate the two profiles, over 2{,}000
splits. The median correlation is 0.92 at 0.75B (5th to 95th percentile 0.79 to
0.99), 0.81 at 1.9B (0.50 to 0.97), 0.93 at 3.7B (0.78 to 0.99), and 0.96 at
4.7B (0.86 to 1.00). The ordering is therefore unlikely to be an artifact of
which eighteen tasks were drawn. The 1.9B interval is the widest, with a 5th
percentile of 0.50, so that rung's ordering is the least stable of the four.

\emph{Parallel forms.} The held-out families mirror the development families
structurally. Whether they also mirror difficulty is an empirical question. On
the same model, across 18 family pairs, the correlation between development and
held-out task difficulty is $+0.23$ $(p=0.36)$, and between their discriminating
power $+0.20$ $(p=0.43)$. With eighteen pairs the interval around each
correlation is wide, roughly $\pm0.45$. A structurally mirrored family therefore
provides no evidence of parallel-form transfer, which is weaker than evidence of
absence. We use the match for coverage of the same capability families, not for
calibration.

\emph{Check-level reach.} The suite has 115 distinct (task, artifact, check)
keys over 117 check entries. Of these, two are passed by every arm, one is
failed by every arm, and the remaining 112 separate at least two arms. The
check that no arm passes is a word floor on HWF-09's benchmark report, which is
met in none of the fifteen cells that produce the file, so it is unreachable at
this scale rather than informative. The two that carry no information are a
substring check on HWF-03's script and a compile check on one HLH-03 module.

\subsection{The mechanism gradient}
\label{sec:mechanism-design}

The matrix shows where the harnesses separate. The gradient asks a narrower
question: whether the completion-control family carries \system{}'s own
advantage, and whether that contribution changes with size. The three-arm
completion-control ablation is repeated at 0.75B, 1.9B, 3.7B and 4.7B, eighteen
tasks per arm. \emph{Full} runs all six executable checks. \emph{Guards off}
disables all six and keeps the todo-sync reminder and the textual re-read.
\emph{Text-only} disables the six checks and the todo-sync reminder and keeps
the textual re-read alone.

Two limits follow from the design. The gradient ablates one harness, so it cannot
explain the span across harnesses, which is carried by arms other than
\system{}. The third rung, 3.7B, runs on granite, so it is the ladder's only
cross-family point and is reported as a check rather than entered into the
slope, even though the gap to the comparator is smallest there. If the guard
contribution is flat across the three same-family rungs, then whatever moves
\system{}'s advantage with size lies outside this family.

\subsection{Analysis plan}
\label{sec:analysis}

The unit of analysis is the task. Comparisons use paired task-level bootstrap
intervals (10{,}000 resamples, seed 42; the replication checks of
Section~\ref{sec:contrasts} resample 4{,}000 times with seed 20261115), the
paired sign proportion, and Wilcoxon
signed-rank tests. Zero differences are discarded, as the signed-rank test
requires, and the $p$ values come from the normal approximation with the tie
correction, not from the exact distribution, because the absolute
differences are heavily tied within a rung. At 4.7B eight of the eighteen tasks are
exact ties, and those pairs drop out of the signed-rank test. Slope intervals
resample the 18 tasks inside each rung and refit on $\log_{10}$ size.

The comparator for the confirmatory contrasts is goose, fixed before this ladder
ran because it was the strongest baseline in the earlier matrix
\citep{mingbird2026}, and not chosen from this ladder's results.
The confirmatory family has seven tests: the slope of the five-arm span, a shape
test on the gap, the slope of the guard contribution, and four per-rung
\system{}-versus-goose contrasts, with Holm correction. The 35.5B anchor is
analyzed separately and is not part of the family. Everything else in the paper
is descriptive.

The hypotheses and their estimators were fixed before the ladder's second rung
started. The four per-rung contrasts reported here are the ones the family
contains. The rescored reading is a second reading of the same cells, applied
where a run reached the wall budget with at least one declared artifact on disk.
It is not a pre-specified test, and where the two readings disagree the paper
reports both.

The minimum detectable effect is not small. With eighteen paired tasks and the
observed paired standard deviation, which runs from 0.34 at 4.7B to 0.58 at
3.7B, the design detects a difference of 0.22 to 0.39 at 80\% power. The value
is reported next to each gap. Repeated collection of one configuration moves an
arm mean by up to 0.233 (the matrix and gradient collections differ by 0.233,
0.229, 0.148 and 0.012 at the four rungs, and the two 3.7B gradient collections
differ by 0.066), so absolute levels are read with that scale in mind. Paired
contrasts are computed within a collection and are not compared across
collections. The design is built to detect large effects: a harness that
completes a quarter of the tasks and one that completes three quarters.

The scorer cannot tell a document built from the fixture from a plausible one
the model invented, so we audit that separately. For every cell whose task has
identifier-bearing fixtures, we compute the fraction of record ids and DOIs in
its artifacts that appear verbatim in the task's fixture file as shipped with the
suite, and report the complement as the fabricated-identifier rate. The audit
does not enter the score. It tells the reader whether a high score came with the
right content.

\section{Results}
\label{sec:results}

Four research questions organize the results. RQ1 asks whether the
completion-controlling harness scores above the third-party harnesses at every
size. RQ2 asks whether its advantage changes with size. RQ3 asks what part of
the harness carries the advantage. RQ4 asks how small models fail. A fifth
subsection narrows a caveat that the threats section states rather than answers:
the ladder holds the harnesses fixed but not the bit width, and a compression
that moves the two arms unequally would be recorded here as an effect of size.

\subsection{Score by size}
\label{sec:levels}

\begin{figure}[t]
\centering
\includegraphics[width=\linewidth]{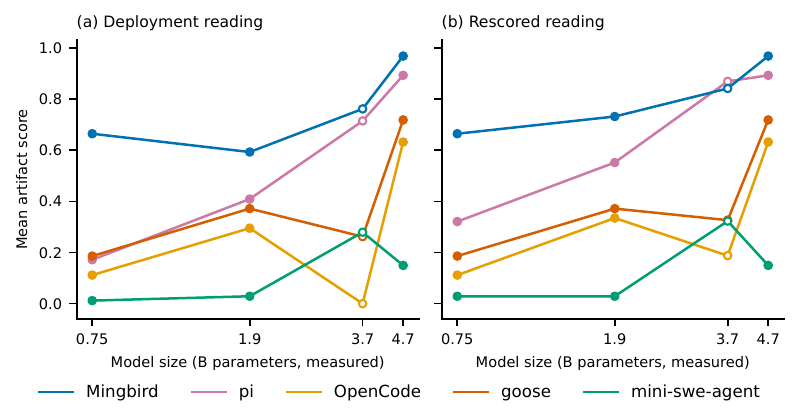}
\caption{Mean artifact score per harness across the ladder, under the two
readings of Table~\ref{tab:ladder}. The horizontal axis is the measured
parameter count on a log scale. Filled markers are the three Qwen3.5 rungs and
open markers the 3.7B Granite rung, which is the single cross-family point, so a
rise into or out of that rung carries a change of family as well as a change of
size. Panel (b) differs from panel (a) only in the 58 cells that reached the
wall budget with declared artifacts on disk. The largest move is pi at 3.7B,
from 0.714 to 0.869, and \system{} moves at 1.9B and 3.7B by 0.139 and 0.079.}
\label{fig:ladder}
\end{figure}

Table~\ref{tab:ladder} reports the mean artifact score for every harness and
model, and Figure~\ref{fig:ladder} plots the same means against the measured
parameter count. An arm's mean is taken over the usable cells it has, which is
17 at 0.75B for \system{} and goose and 17 at 3.7B for \system{}, and 18
everywhere else.

The ranking is \system{} first at every rung. At 0.75B the five separate by
0.653, and \system{} produces every declared artifact on eight of its seventeen
tasks against three for pi, two for goose and none for the two smallest
harnesses. At 1.9B it leads pi, goose, OpenCode and mini-swe-agent in that order,
and completion comes apart from score, since every artifact is present on twelve
of its eighteen cells against six for pi and none for mini-swe-agent. At 3.7B
pi draws level at the top, and the order below it reshuffles: mini-swe-agent
moves from fifth to third while OpenCode falls to last. Artifact production does
not track the ranking there either: pi completes fifteen full sets to
\system{}'s fifteen,
and two of pi's complete sets reached the wall budget, so a full set of files and
a zero coexist. The paired contrast at that rung is 0.005, and with the pair at
1.9B it is one of only two in the table whose interval includes zero under both
readings. At 4.7B every declared artifact is present on seventeen of eighteen
\system{} cells, against thirteen for pi and ten for goose.

The ladder is read across the two axes at once. \system{} at 0.75B averages
0.664 in the primary collection (17 usable cells), 0.503 in a repeat (14) and
0.431 in the mechanism gradient's full arm (18), which repeats the same
configuration, so its own level moves by 0.23 with nothing changed. The three
cell sets differ because only the ladder and its repeat carry the exclusion
rule, so these are three measurements of one condition rather than three draws
from one sample. Across the whole 6.3-fold range on the card
the winning harness gains 0.304. The other four gain between 0.14 and 0.72: pi
0.72, goose 0.53, OpenCode 0.52 and mini-swe-agent 0.14, the last from a floor
where its own reply contract rejects the model rather than from a graded
capability. The winner's gain is small because it starts at 0.664 with a third
of the scale left, while pi starts at 0.172 with four times the room. The
five-arm spread at a fixed size is 0.564 to 0.819.

\begin{table}[t]
\centering
\small
\setlength{\tabcolsep}{4pt}
\caption{Mean artifact score per model and harness (0--1). The upper panel is the deployment reading, in which a cell that exhausts its wall budget scores zero. The lower panel rescores the archived working directory of the 58 cells that reached the budget with declared artifacts already on disk, using the same checker. The two panels differ only in those cells, 36 of which are at 3.7B. An arm's mean is taken over the usable cells it has: \system{} has 17 at 0.75B and 3.7B and goose 17 at 0.75B, because three infrastructure cells are excluded by the rule in Section~\ref{sec:threats}. Span is best minus worst across all five harnesses at that rung.}
\label{tab:ladder}
\begin{tabular}{@{}lccccc c@{}}
\toprule
Model (B) & \system{} & pi & OpenCode & goose & mini & Span \\
\midrule
\multicolumn{7}{@{}l}{\emph{Deployment reading: a cell at the wall budget scores zero}}\\
0.75 & 0.664 & 0.172 & 0.112 & 0.186 & 0.012 & 0.653 \\
1.9 & 0.593 & 0.409 & 0.295 & 0.372 & 0.029 & 0.564 \\
3.7 & 0.762 & 0.714 & 0.000 & 0.262 & 0.279 & 0.762 \\
4.7 & 0.968 & 0.893 & 0.632 & 0.718 & 0.150 & 0.819 \\
\midrule
\multicolumn{7}{@{}l}{\emph{Rescored reading: the archived directory of a timed-out cell is scored}}\\
0.75 & 0.664 & 0.321 & 0.112 & 0.186 & 0.029 & 0.636 \\
1.9 & 0.732 & 0.552 & 0.335 & 0.372 & 0.029 & 0.703 \\
3.7 & 0.841 & 0.869 & 0.188 & 0.327 & 0.323 & 0.681 \\
4.7 & 0.968 & 0.893 & 0.632 & 0.718 & 0.150 & 0.819 \\
\bottomrule
\end{tabular}
\end{table}

\subsection{Every harness at every size}
\label{sec:contrasts}

Table~\ref{tab:contrast} reports the paired contrast between \system{} and each
of the four third-party harnesses at each rung, under both readings of the data.
The two readings answer different questions. Under the deployment reading a cell
that exhausts its wall budget scores zero, which is what a machine with a fixed
overnight schedule experiences. That zero conflates a harness that could not do
the task with a harness that did the task but not inside the budget. The second
reading takes the archived working directory of each of the 58 cells that ended
at the wall budget with artifacts already on disk and rescores it with the same
checker. It asks whether the work was there, and it says nothing about whether
the work arrived in time.

Thirty-six of the 58 rescored cells are at 3.7B, so the second reading moves that
rung more than any other. pi rises there from 0.714 to 0.869 and \system{} from
0.762 to 0.841, which turns the pair's 0.005 into $-0.020$. goose rises from
0.262 to 0.327, OpenCode from 0.000 to 0.188, and mini-swe-agent from 0.279 to
0.323. At 1.9B the reading lifts pi from 0.409 to 0.552 and \system{} from
0.593 to 0.732, while goose does not move at 0.372. The two arms move together
often enough that the contrast survives.

Under the deployment reading 13 of the sixteen intervals exclude zero. The
smallest is the contrast with pi at 4.7B, $+0.075$ with a 95\% interval of
$[+0.012,+0.147]$, and the largest is the contrast with mini-swe-agent at the
same rung, $+0.819$ $[+0.647,+0.964]$. The three that include zero are all
against harnesses that keep their strength at that size: pi at 1.9B ($+0.184$,
$[-0.080,+0.442]$) and at 3.7B ($+0.005$, $[-0.230,+0.241]$), and goose at 1.9B
($+0.221$, $[-0.002,+0.436]$). Under the rescored reading 14 of the sixteen
exclude zero, and the exceptions are the pairs with pi at 1.9B and 3.7B.

The 3.7B exception has a cost signature. pi's
mean cell at 3.7B takes 8.1 minutes against 5.3 for \system{}, and its artifact
sets there are completed later rather than not at all. A practitioner choosing
between the two at that size is trading a score difference that is not
distinguishable from zero against a half again as long a run. Under the
deployment reading, which is the one that matches a fixed schedule, the slower
harness pays for that trade.

The four contrasts against goose are the pre-specified comparisons and form part
of a seven-test confirmatory family with Holm correction
(\mbox{Section~\ref{sec:analysis}}). Two survive the correction: the contrast at 0.75B
($p=0.0010$, adjusted $0.0067$) and the one at 4.7B ($p=0.0068$, adjusted
$0.0410$). The 3.7B contrast has a raw $p=0.016$ and an adjusted value of 0.080,
and the 1.9B contrast does not approach the threshold. An exact sign-flip
permutation over the same pairs, which assumes nothing about the shape of the
difference distribution, gives $p=0.0001$, $0.0784$, $0.0068$ and $0.0039$: the
two survivors keep their place, and the 3.7B contrast joins them at an adjusted
0.034. The signed-rank test is the one registered in advance, so the
confirmatory reading stays at two, and the 3.7B rung is reported as boundary
evidence rather than as a third confirmed result. The other twelve
contrasts in Table~\ref{tab:contrast} use the same interval procedure and are
reported as descriptive, because they were not among the four registered in
advance rather than because anything about their baselines was chosen after the
fact. The winner's-curse argument belongs to the gap-to-best estimator of
Section~\ref{sec:shape}, which picks its comparator from this ladder's results.
The four baselines here are a fixed set.

\begin{table}[t]
\centering
\small
\setlength{\tabcolsep}{4pt}
\caption{Paired contrast between \system{} and each established harness at each rung: mean difference in artifact score with a task-level bootstrap 95\% interval over 10{,}000 resamples. A pair uses the tasks where both arms have a usable cell. The deployment reading scores a cell that exhausts its wall budget as zero. The rescored reading replaces the zero of a timed-out cell with the score of its archived working directory. An interval that excludes zero is printed in bold. The four goose rows are the pre-specified comparisons and belong to the seven-test Holm-corrected family. The other twelve were not registered in advance and are descriptive.}
\label{tab:contrast}
\setlength{\tabcolsep}{3pt}
\begin{tabular}{@{}llcc@{}}
\toprule
Model (B) & Baseline & Deployment reading & Rescored reading \\
\midrule
0.75 & pi & \textbf{+0.483 [+0.256,\,+0.681]} & \textbf{+0.325 [+0.102,\,+0.546]} \\
 & OpenCode & \textbf{+0.546 [+0.349,\,+0.731]} & \textbf{+0.546 [+0.349,\,+0.731]} \\
 & goose & \textbf{+0.508 [+0.339,\,+0.682]} & \textbf{+0.508 [+0.334,\,+0.679]} \\
 & mini & \textbf{+0.652 [+0.483,\,+0.812]} & \textbf{+0.634 [+0.466,\,+0.790]} \\
\midrule
1.9 & pi & +0.184 [$-$0.080,\,+0.442] & +0.180 [$-$0.006,\,+0.371] \\
 & OpenCode & \textbf{+0.298 [+0.063,\,+0.528]} & \textbf{+0.397 [+0.158,\,+0.605]} \\
 & goose & +0.221 [$-$0.002,\,+0.436] & \textbf{+0.360 [+0.131,\,+0.566]} \\
 & mini & \textbf{+0.564 [+0.361,\,+0.758]} & \textbf{+0.703 [+0.545,\,+0.844]} \\
\midrule
3.7 & pi & +0.005 [$-$0.230,\,+0.241] & $-$0.020 [$-$0.199,\,+0.162] \\
 & OpenCode & \textbf{+0.762 [+0.572,\,+0.924]} & \textbf{+0.643 [+0.370,\,+0.860]} \\
 & goose & \textbf{+0.484 [+0.199,\,+0.729]} & \textbf{+0.496 [+0.229,\,+0.724]} \\
 & mini & \textbf{+0.466 [+0.267,\,+0.661]} & \textbf{+0.499 [+0.295,\,+0.685]} \\
\midrule
4.7 & pi & \textbf{+0.075 [+0.012,\,+0.147]} & \textbf{+0.075 [+0.012,\,+0.147]} \\
 & OpenCode & \textbf{+0.336 [+0.155,\,+0.524]} & \textbf{+0.336 [+0.163,\,+0.523]} \\
 & goose & \textbf{+0.250 [+0.107,\,+0.408]} & \textbf{+0.250 [+0.111,\,+0.408]} \\
 & mini & \textbf{+0.819 [+0.647,\,+0.964]} & \textbf{+0.819 [+0.652,\,+0.956]} \\
\bottomrule
\end{tabular}
\end{table}

Task by task, \system{} beats goose on 14, 11 and 11 of the comparable tasks at
0.75B, 1.9B and 3.7B, and loses on 0, 3 and 1, so the mean gap is not the work
of a few large wins. At 4.7B the two arms tie on eight tasks, every one of them
at 1.000. \system{} wins nine and loses one, HWF-13, where goose reaches 1.000
and \system{} 0.857. The suite has saturated for this pair at the top of the
ladder, which is the ceiling discussed in Section~\ref{sec:shape}.

Two rungs, 0.75B and 4.7B, were collected a second time, which repeats the
decisive contrasts on the same tasks (72 cells, of which the same infrastructure
rule removes eight). At 4.7B the repeat is nearly exact. \system{} scores 0.968
and then 0.988, goose 0.718 in the matrix and 0.731 in the repeat over the
seventeen tasks it has in both. The paired gap is $+0.250$ $[+0.107,+0.408]$
over the eighteen tasks both arms complete in the matrix and $+0.256$
$[+0.105,+0.437]$ over the seventeen in the repeat, and fifteen of the eighteen
\system{} cells repeat their score to the digit. At 0.75B the level
falls and the gap does not. On the ten tasks where both arms are clean in both
collections \system{} drops 0.156 and goose 0.01, and the paired gap is $+0.420$
and then $+0.423$. Over every usable cell \system{} falls from 0.664 to 0.503
and goose from 0.186 to 0.154. The paired gap over the tasks both arms complete
is $+0.508$ $[+0.339,+0.682]$ and then $+0.476$ $[+0.260,+0.701]$, fourteen wins
and no losses the first time and nine and none the second. Those counts are
over usable cells, and the exclusions matter to them: restoring the nine cells
the rule removes at this rung, two in the matrix and seven in the repeat, gives
fifteen wins and one loss in the matrix and twelve
wins and two losses in the repeat, $+0.466$ $[+0.249,+0.664]$ and $+0.315$
$[+0.061,+0.541]$, because every one of the three losses is a cell on which one
arm's run was cut short by the server rather than by the model. Under both exclusion rules the
contrast stays positive while its size moves. The level at the smallest rung is
a draw from a wide distribution, the contrast is not.

The winner is not the arm that moves most. The precision ladder runs the
published granite tag again as its unquantized level, which repeats the 3.7B
cells for \system{} and pi in a collection that ran at the matrix's speed. pi
falls from 0.714 to 0.493 there, a change larger than the winner's at 0.75B,
while \system{} moves from 0.762 to 0.822 over the same eighteen cells. The repeats do not support a stable level for any arm,
\system{} included. They support a paired contrast that survives where both
arms are measured again.

\begin{figure}[t]
\centering
\includegraphics[width=\linewidth]{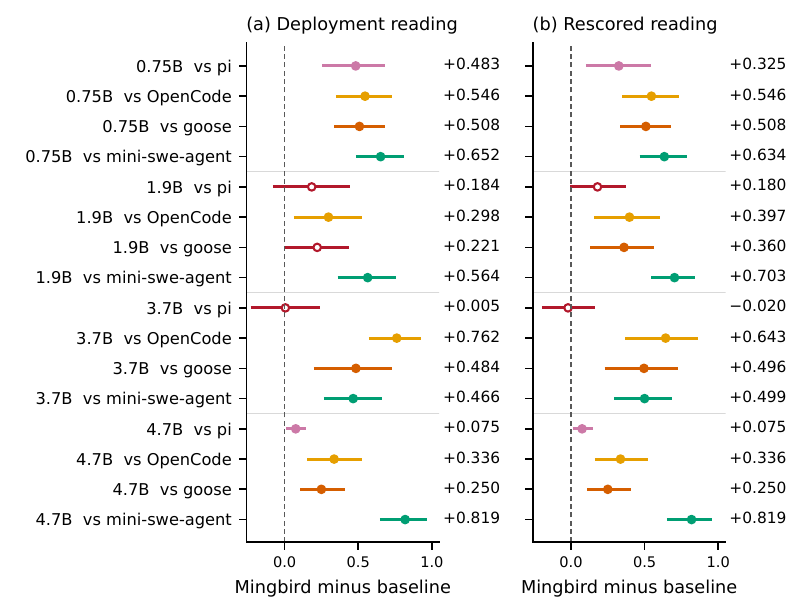}
\caption{The sixteen paired contrasts of Table~\ref{tab:contrast} with their
task-bootstrap intervals, read from the same payload as the table. A filled
marker means the interval excludes zero. An open marker means it includes zero,
and a marker is drawn in red when its interval crosses zero and in the arm's
color otherwise. Under the deployment reading every point estimate lies to the
right of the dashed line at zero, and under the rescored reading all but one do.
The intervals are tightest at 4.7B, where the two arms separate on nearly every
task, and the verdicts that change between readings sit at 3.7B, where 36 of the
58 rescored cells are.}
\label{fig:forest}
\end{figure}

\subsection{How the advantage changes with size}
\label{sec:shape}

\begin{figure}[t]
\centering
\includegraphics[width=\linewidth]{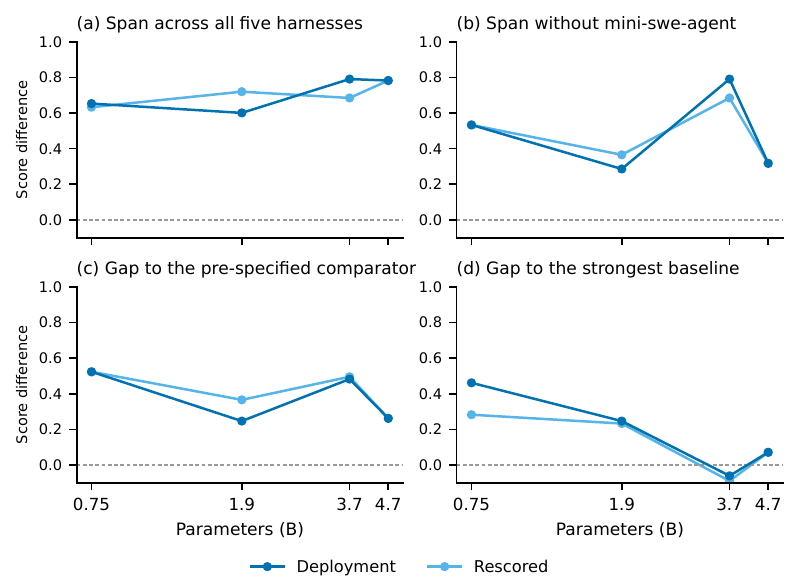}
\caption{Four measures of harness leverage at each rung, computed from the same
357 usable cells under both readings and on the common task panel of
Table~\ref{tab:span}. The five-arm span in (a) is 0.653, 0.601, 0.791 and 0.782,
and its slope is $+0.198$ with an interval that includes zero. Removing the arm
whose floor is a reply contract (b) leaves a spread that shows no trend at all.
The pre-specified gap to goose (c) is non-monotone, and the gap to the best
baseline (d) falls from 0.461 to 0.071 because the strongest competitor changes
from goose to pi. The two readings differ mainly at 3.7B, where most of the
rescored cells sit.}
\label{fig:span}
\end{figure}

The next question is whether the harness matters more or less as the model
grows. Figure~\ref{fig:span} and Table~\ref{tab:span} answer it with four
estimators on one common task panel: the span across five harnesses, the span
with the reply-contract arm removed, the gap to the pre-specified comparator,
and the gap to whichever baseline is strongest at that rung.

On that panel the spans are 0.653, 0.601, 0.791 and 0.782 at the four rungs.
They differ by up to 0.04 from Table~\ref{tab:ladder} because the panel
drops the three tasks that one arm is missing on. Their slope is $+0.198$ with
an interval of $[-0.078,+0.476]$, consistent with a rise of half a point per
decade and with a fall of a tenth. The four-arm span, which
drops the reply-contract floor, has no slope either ($+0.011$,
$[-0.302,+0.257]$), and the pre-specified gap to goose does not resolve across
the four rungs ($-0.195$, $[-0.494,+0.100]$).

Within the Qwen family alone that gap does resolve, at $-0.332$ with an interval
of $[-0.639,-0.025]$, and it is the one place in the paper where a
within-family trend clears its interval. It is a level difference, not a trend:
the three points are 0.508, 0.231 and 0.245, so the whole decline is
the first step, off the rung where two baselines leave the task empty, and the
last step goes the other way. A line through those three points is not a trend
estimate, and the quadratic test does not prefer a curve to a line ($p=0.572$),
so neither shape is identified, and the estimate is reported in the table for
completeness rather than as a slope.

One quantity does move. The margin over the best baseline at each rung is 0.461,
0.246, $-0.061$ and 0.071, a slope of $-0.591$ $[-0.836,-0.251]$. The decline
measures the baselines rather than the harness. The best baseline at 0.75B is
goose at 0.186. At the other three rungs it is pi. pi climbs from 0.172 to 0.893
across the ladder, a steeper trajectory than \system{}'s 0.664 to 0.968, so the
leader's margin narrows while the spread among harnesses stays the same. A study
that compared the winner only against the current champion would read that as
leverage shrinking with scale. The spans, which hold the comparison set fixed,
do not.

Two properties of the design set what the ends of the ladder can show. At 3.7B
the cross-family model lets pi draw level: the gap to best is $-0.061$ on the
common panel and the paired interval includes zero, so this rung is a tie at
both readings. At 4.7B the suite saturates: on the full eighteen tasks,
\system{} and goose both reach 1.000 on eight of them, and on the table's
fifteen-task panel the count is six, where the goose gap is 0.262 against 0.436
on the nine tasks the pair does not both solve. Both are properties of the
measurement rather than of the harnesses.

Taken together, the three artifacts give the ladder a specific limit. A floor arm
that fails at every size manufactures a rise in the span. A ceiling at the top
manufactures a decline in the winner's margin. A single cross-family rung in the
middle manufactures a peak. Each of the three has been read as a property of
harness leverage, and each is reproducible here by choosing the estimator. The
pattern is not specific to harnesses: whether a capability trend appears at all
depends on the metric chosen to look for it \citep{schaeffer2023mirage}. What
the ladder does establish is the level: at every rung the harnesses are separated
by a margin the intervals resolve. What it cannot establish is the slope, and no
arrangement of these four rungs would. A leverage-versus-size claim needs three
things this ladder does not have: a baseline set with no floor arm, tasks that
stay unsaturated at the top rung, and at least two model families at every rung,
or a within-family slope alone.

\begin{table}[t]
\centering
\small
\setlength{\tabcolsep}{4pt}
\caption{Four candidate measures of harness leverage at each rung, under the deployment reading, and the slope of each on $\log_{10}$ parameter count. \emph{Span} is best minus worst across all five harnesses. \emph{Span$_{-\mathrm{mini}}$} drops mini-swe-agent, whose floor is its own reply-format contract rather than a graded capability difference. \emph{Gap to goose} is the pre-specified comparator. \emph{Gap to best} uses whichever baseline scored highest at that rung, a selection made from these data. \emph{Saturated} counts tasks on which \system{} and goose both reach 1.000, and the last column recomputes the goose gap on the remaining tasks. Every value in a row is taken on one common task panel (15 tasks for the four-rung panel, 16 for the three Qwen3.5 rungs), so the same tasks are read at every rung. The level columns therefore differ in the last digits from Table~\ref{tab:ladder}, whose means use each arm's own usable cells. A slope resamples that panel, so the comparison across rungs is paired.}
\label{tab:span}
\footnotesize
\setlength{\tabcolsep}{3.5pt}
\begin{tabular}{@{}lcccccc@{}}
\toprule
Model (B) & Span & Span$_{-\mathrm{mini}}$ & Goose gap & Best gap & Saturated & Unsat.\ gap \\
\midrule
0.75 & 0.653 & 0.533 & +0.523 & +0.461 & 0 of 15 & +0.523 \\
1.9 & 0.601 & 0.286 & +0.246 & +0.246 & 2 of 15 & +0.284 \\
3.7 & 0.791 & 0.791 & +0.482 & $-$0.061 & 1 of 15 & +0.517 \\
4.7 & 0.782 & 0.318 & +0.262 & +0.071 & 6 of 15 & +0.436 \\
\bottomrule
\end{tabular}

\vspace{0.6em}
\footnotesize
\setlength{\tabcolsep}{3pt}
\begin{tabular}{@{}lcc@{}}
\toprule
 & \multicolumn{2}{c}{Slope on $\log_{10}$ size [95\% interval]} \\
\cmidrule(lr){2-3}
Quantity & All four rungs & Qwen3.5 rungs only \\
\midrule
Span & +0.198 [$-$0.078,\,+0.476] & +0.206 [$-$0.132,\,+0.507] \\
Span$_{-\mathrm{mini}}$ & +0.011 [$-$0.302,\,+0.257] & $-$0.255 [$-$0.579,\,+0.031] \\
Gap to goose & $-$0.195 [$-$0.494,\,+0.100] & \textbf{$-$0.332 [$-$0.639,\,$-$0.025]} \\
Gap to best & \textbf{$-$0.591 [$-$0.836,\,$-$0.251]} & \textbf{$-$0.459 [$-$0.701,\,$-$0.151]} \\
\bottomrule
\end{tabular}
\end{table}

\subsection{What carries the advantage}
\label{sec:mechanism-results}

The mechanism gradient repeats one harness with its completion controls switched
off, at 0.75B, 1.9B, 3.7B and 4.7B, eighteen tasks per arm.
Table~\ref{tab:bgradient} reports the three arms at each rung.

At 0.75B the full gate averages 0.431 and disabling the six executable checks
costs 0.080, with a paired interval of $[-0.183,+0.320]$ that includes zero. At
1.9B the same contrast is 0.117 $[-0.047,+0.302]$. At 4.7B it is $-0.004$
$[-0.052,+0.044]$: full 0.980, guards off 0.984. The ablation also runs at 3.7B,
on granite, where the difference is $+0.003$ $[-0.032,+0.038]$.

Set against the ladder, those numbers are small. The lead over goose is 0.508 at
0.75B, 0.221 at 1.9B, 0.484 at 3.7B and 0.250 at 4.7B. The point estimate of the
gate's contribution is therefore 16\% of the lead at 0.75B, 53\% at 1.9B, and
under 2\% at the two rungs above. The intervals bound the contribution tightly at the top:
at 4.7B the upper limit is 0.044, or 18\% of the lead. At 0.75B the upper limit
is 0.320, so the interval cannot exclude a gate contribution of up to 63\%. The
point estimate is what places it near a sixth.

The failure coding shows where the gate does act. At 1.9B the full arm leaves no
artifact on 0 of 18 tasks, the guards-off arm on 3, and the text-only arm on 1,
so switching the checks off produces three empty runs that the full harness
prevents. At 0.75B the count is 7, 6 and 6: there the checks do not change how
often a run ends empty, and the 0.080 appears in partial scores instead. At
3.7B and 4.7B the counts are 1, 1, 1 and 0, 0, 0. Completion control does what
it was designed to do at 1.9B. At 0.75B the gate is not what produces the 0.508
lead: its point estimate is a sixth of that and the interval admits up to 63\%,
and the empty-run count, which does not rise when the checks are switched off,
is what places it outside the story there.

Where the rest of the lead comes from can be bounded from the other side. Split
each paired gap by whether the baseline left any of its declared artifacts. At
0.75B, 8 of the 17 comparable pi cells are empty and they carry 51\% of the gap
to pi, 10 of 16 goose cells 68\%, and 11 of 17 OpenCode cells 72\%. The
remainder is a difference on cells both arms completed, and it is large:
\system{} is $+0.443$ $[+0.085,+0.714]$ ahead of pi and $+0.439$
$[+0.248,+0.628]$ ahead of goose on the tasks those arms did produce. At 1.9B
the balance shifts almost entirely to availability: the empty cells carry 94\%
of the gap to pi and 90\% of the gap to goose, and on produced cells the gap is
$+0.016$ $[-0.261,+0.313]$ and $+0.032$ $[-0.210,+0.246]$. At 3.7B the pair with
pi is level on the 16 tasks both arms complete ($-0.057$, $[-0.267,+0.146]$). At
4.7B the character reverses: pi leaves no empty cell, and the whole $+0.075$
$[+0.012,+0.147]$ lead is a difference in how much of the task the two arms
finish.

The advantage is therefore a progress effect at every rung: at the small end it
is mostly a difference in whether a run gets anywhere, and at the top it is a
difference in how far a completed run gets. The six checks produce neither. What
must produce it is whatever keeps a weak model working after a tool error or
inside a repetition loop, and the transcripts name one such process exactly.

At 0.75B, eleven of pi's eighteen cells contain the shell's answer to an
unquoted Unix-rooted path, ``cd: too many arguments'', and nine of those eleven
end with nothing written. The tool surface explains where the command comes
from: pi exposes read, bash, edit and write, so a model that wants to list a
directory has to do it through the shell, and the shell is where a Unix-rooted
path with a space fails. In HWF-02 the model then issues a directory-listing
\texttt{bash} call 274 times, 273 of them byte-identical: the path fails on
every turn, the error is appended to the context, and the request grows by 113
tokens a turn from 1,844 to 32,767, where the run stops. The same model on the same task in \system{} scores 0.786, and
\system{}'s tools take the directory as a native call rather than a shell
command. Its transcript records the repeated call disabled, ten further
attempts, a context reset from 7,033 to 2,237 tokens, and then the artifact
written and the task finished. Across the rung no \system{} cell exceeds 29
identical calls and the intercept fires in all seventeen. Pi has twelve cells
above 100 identical calls, goose eight and OpenCode four.

That is the process the decomposition's empty-cell channel is made of: a single
failing action repeated until the window fills. A harness that interrupts the
repeat and resets the context converts the run. One that does not reaches the
iteration cap with an empty directory. The completion gate sits outside this
chain, which is why switching it off changes the empty-run count at 0.75B by one
cell in the wrong direction.

One rule in that chain can be switched off alone. The build's
\texttt{anti\_loop} ablation disables the streak rule, which blocks the same
tool call eight times in a row and the same call signature four times, and
leaves the read-loop intercept and the hard context reset in place. Eighteen
tasks were run twice at 0.75B, full and streak-off, interleaved inside one
batch. The full arm scores 0.411 higher on average, $[+0.218,+0.608]$, higher on
twelve tasks and lower on one. On eight tasks it is the only arm that writes any
declared artifact, and the reverse never occurs. The switch is visible in the
transcripts: the full arm's eighteen cells record 253 calls to a tool the
harness had already disabled, and the off arm's record 55. The 55 are 45 todo
calls and 10 web calls, and no file tool is among them, because what survives
the switch belongs to the todo and web rules it does not touch. The point
estimate sits close to what a rerun moves a Mingbird cell by at this rung (0.37
on average), and one run per cell cannot separate a 0.411 mean from that noise.
The artifact split is a separate observation, and it is the one repeated runs
would have to close: eight tasks produce a declared artifact in the full arm
and none in the off arm, no task goes the other way, and a two-sided sign test
over those eight disagreements gives $p=0.0078$, conditional on the eight and
treating tasks as independent.

The process changes with rung, and the counts say how. At 1.9B none of pi's five
empty cells ends normally: four reach the wall budget and one stalls, so the
availability channel there is time rather than repetition. Rescoring those
cells' working directories lifts pi by 0.143 and \system{} by 0.139, which
leaves that rung's contrast at $+0.180$ $[-0.006,+0.371]$: the gap is not an
artifact of scoring a timeout as zero. At 3.7B nine of OpenCode's twelve empty
cells and eight of goose's nine are budget timeouts. At 4.7B mini-swe-agent
accounts for 13 of the 16 empty cells, and ten of its thirteen end normally on
its reply contract.

The todo-sync reminder is the one intervention switched off in the text-only arm
and on in both others, so it owns the second contrast. It moves without a
consistent sign: $-0.035$ at 0.75B, $-0.079$ at 1.9B, $+0.095$ at 3.7B and
$-0.016$ at 4.7B. The arms reorder as well: the text-only arm is the lowest at
3.7B, 0.812 against 0.910 for the full gate, and the highest at 4.7B, 1.000
against 0.980. The 3.7B difference is a level in one collection, and the repeat
collection does not reproduce it.

The slope of the guard contribution on $\log_{10}$ size is $-0.104$ with an
interval of $[-0.414,+0.220]$ and a bootstrap $p=0.525$. The contribution grows
or shrinks with size by no amount this design can resolve: it is small and
positive at the two small rungs, zero at the two large ones, and the family's
adjusted $p$ for the slope is 1.000.

The gradient's full arm is a second collection of the matrix's \system{} cell,
which measures how far an arm mean moves between collections. The two agree at
4.7B (0.980 against 0.968) and diverge at the small rungs: 0.431 against 0.664
at 0.75B, 0.822 against 0.593 at 1.9B, and 0.910 against 0.762 at 3.7B. The
spread is bimodal, the same shape the 72-cell replication shows: most tasks
repeat their score exactly and a few move by most of the scale. The gradient
measures one arm, not a contrast, so it cannot carry an ordering claim across
collections. It carries the size of the drift: 0.229 at 1.9B between two
collections of the same configuration, which is the scale against which every
level in this paper is read. The ordering claim rests instead on the matrix and
on the repeat that contains both arms, in Section~\ref{sec:contrasts}.

The 3.7B rung is where that instability is largest in absolute terms. A repeat
of the whole 3.7B gradient puts the full arm at 0.844 against 0.910, and the
guard contrast at $-0.020$ against $+0.003$. Across the two collections the
contrast is zero either way. At the three Qwen rungs the gradient's contrasts
are paired within one collection and never compared across collections, and they
carry intervals because a difference of that size is near what eighteen paired
tasks can resolve.

\begin{table}[t]
\centering
\small
\setlength{\tabcolsep}{2.5pt}
\caption{The mechanism gradient: the same three-arm completion-control ablation
repeated at four rungs, eighteen tasks per arm. All three arms keep the
textual re-read of the task. \emph{Guards off} additionally disables the six
executable checks: the no-work refusal, the test guard, the finish-claim artifact
check, the plan-completeness gate, the artifact-content check, and the post-finish
summary request, while \emph{text only} is the full gate with those six and
the todo-sync reminder disabled. \emph{Guards} is the paired difference between
the full arm and the guards-off arm, with its task-level bootstrap interval and
Wilcoxon $p$. \emph{Reminder} is the difference between the two ablated arms,
which isolates the todo-sync reminder and is reported as a point estimate. The
hypothesis this ablation tests is decided by the slope of \emph{Guards} on
$\log_{10}$ size, not by the three per-rung values. $^{\dagger}$The 3.7B rung runs on granite, the only rung of the
ladder whose model family differs from the rest, so it is reported as a
cross-family check and does not enter the slope. A second collection of the whole
3.7B gradient puts the full arm at 0.844 and this contrast at $-0.020$. The rung
is reported as a level and carries no mechanism claim.}
\label{tab:bgradient}
\begin{tabular}{@{}lcccccc@{}}
\toprule
Model (B) & Full & Guards off & Text only & Guards [95\% CI] & $p$ & Reminder \\
\midrule
0.75 & 0.431 & 0.351 & 0.386 & $+0.080$ [$-0.18$, $+0.32$] & 0.700 & $-0.035$ \\
1.9 & 0.822 & 0.705 & 0.784 & $+0.117$ [$-0.05$, $+0.30$] & 0.241 & $-0.079$ \\
3.7$^{\dagger}$ & 0.910 & 0.907 & 0.812 & $+0.003$ [$-0.03$, $+0.04$] & 0.833 & $+0.095$ \\
4.7 & 0.980 & 0.984 & 1.000 & $-0.004$ [$-0.05$, $+0.04$] & 0.785 & $-0.016$ \\
\bottomrule
\end{tabular}
\end{table}

\subsection{How small models fail}
\label{sec:failure-results}

Each cell is coded on two axes, from its working directory and its runner
record rather than from the transcript. The artifact axis records whether all,
some, or none of the declared artifacts exist. The termination axis records how
the run ended: finished, wall budget, a no-artifact exit in which the harness
reported completion with nothing written, stall, early finish, or an error exit
in which the harness process returned nonzero. The two axes are independent and
keeping them apart matters, because a cell can finish with no artifacts and a
cell can be stopped with all of them. Cells whose failure is infrastructure
and not agent behavior are excluded by the rule in
Section~\ref{sec:threats} and re-run.

\begin{figure}[t]
\centering
\includegraphics[width=\linewidth]{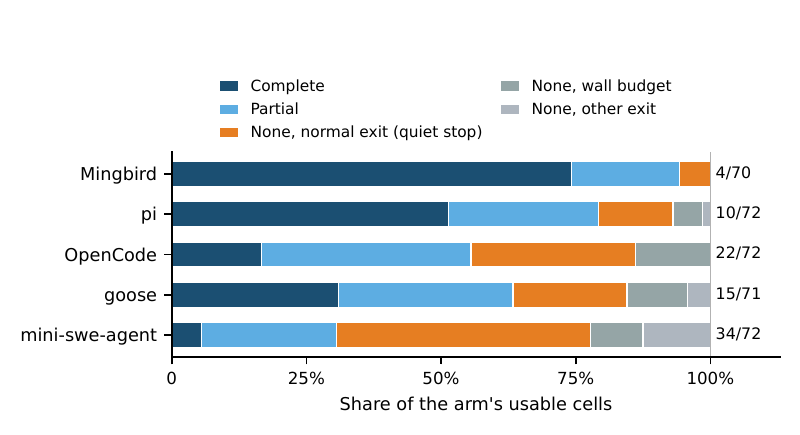}
\caption{Where each arm's cells end, as a share of the usable cells it has. The
orange segment is the quiet stop: a run that exits normally and leaves none of
its declared artifacts. It is 4 of 70 cells for \system{} and 34 of 72 for
mini-swe-agent, with pi at 10 of 72, goose at 15 of 71 and OpenCode at 22 of 72.
The figure reads the same counts as Table~\ref{tab:bfailures}.}
\label{fig:failures}
\end{figure}

Table~\ref{tab:bfailures} and Figure~\ref{fig:failures} aggregate the coding
over the 357 usable ladder cells. Of the 127 cells that produce none of their
declared artifacts, 85 end
with a normal runner exit and 29 hit the wall budget. A harness can therefore
fail a task while its process reports success, which is why the artifact axis is
coded from the working directory rather than from an exit code. The rate ranges
from 4 of 70 cells for \system{} to 50 of 72 for mini-swe-agent, with pi at 15
of 72, goose at 26 of 71 and OpenCode at 32 of 72.

The share is rung-dependent. At 0.75B nearly all artifact-free cells end
normally, 43 of 48. At 1.9B the share falls to 19 of 33, and at 3.7B only 10 of
30 do, because there the no-artifact cells mostly run to the wall budget: 20 of
the 30, of which nine are OpenCode cells and eight goose cells. At 4.7B the
quiet stop returns: 13 of 16 artifact-free cells end normally, and ten of those
13 belong to mini-swe-agent.

The 85 normal exits are not all quiet in the same sense, and a third pass over
the transcripts, separate from the two-axis coding, says which are. A cell is
error-bearing when its transcript records a traceback, an explicit error line, a
repeated-format error or a parse error, with terminal color codes stripped
first. The quiet stop is concentrated in two arms: 14 of pi's 15 artifact-free
cells record no line matching those patterns, and 22 of OpenCode's 32 do not.
Nine of pi's fifteen empty cells and nine of OpenCode's thirty-two carry the
shell's ``cd: too many arguments'' as ordinary tool output, which the patterns do
not treat as an error because the harness passes the message through instead of
raising it, so a clean transcript is not a clean run. What the coding
measures is whether the harness reported a failure, not whether one occurred.
goose sits between them, at 18 of its 26 artifact-free cells recording no error line.
mini-swe-agent is the opposite extreme: 40 of its 50 artifact-free
cells carry an error line, and 34 of them carry its own
\texttt{RepeatedFormatError}, so that arm usually stops on a named contract
rather than silently. \system{} leaves four artifact-free cells, all of which
end normally and all four of which record an error line in the transcript.

The distinction sets what a scoring rule can and cannot substitute for. A rule
that read a normal process exit as completion would credit 85 runs that wrote
none of their declared artifacts, and 40 of them would carry no error line in
the transcript as well. The arms share an exit status that does not record the
failure, rather than a process that fails.

\begin{table}[t]
\centering
\small
\setlength{\tabcolsep}{4pt}
\caption{What each arm leaves on disk, over the 357 usable ladder cells. \emph{Complete} means every declared artifact exists, \emph{partial} that some do, and \emph{none} that the working directory holds none of them. The three right-hand columns split the empty cells by how the run ended: a normal exit, the wall budget, or another termination. The coding reads the working directory and the runner record, not the transcript. Each arm's total is the number of usable cells it has: 70 for \system{} and 71 for goose, after the three infrastructure exclusions.}
\label{tab:bfailures}
\setlength{\tabcolsep}{1.5pt}
\begin{tabular}{@{}lcccccc@{}}
\toprule
Arm & Complete & Partial & None & None: finished & None: budget & None: other \\
\midrule
\system{} & 52 & 14 & 4 & 4 & 0 & 0 \\
pi & 37 & 20 & 15 & 10 & 4 & 1 \\
OpenCode & 12 & 28 & 32 & 22 & 10 & 0 \\
goose & 22 & 23 & 26 & 15 & 8 & 3 \\
mini-swe-agent & 4 & 18 & 50 & 34 & 7 & 9 \\
all & 127 & 103 & 127 & 85 & 29 & 13 \\
\bottomrule
\end{tabular}
\end{table}

Three failure shapes recur. At 0.75B, pi's and OpenCode's zero cells
are dominated by a host mismatch: the model emits an unquoted Unix path, the
shell answers ``cd: too many arguments'', and the working directory stays empty.
pi records that error in eleven of its eighteen cells and OpenCode in six.
Neither \system{} nor goose nor mini-swe-agent records it at that rung, and
OpenCode records it in nine cells across the other three rungs. mini-swe-agent
fails by a second route, because its loop requires a bash block in every reply
and a model that repeats a non-conforming reply ends the run with the harness's
own \texttt{RepeatedFormatError}. That status appears in fifteen of eighteen
cells at 0.75B, fifteen at 1.9B, none at 3.7B and sixteen at 4.7B. At 3.7B the
format error disappears and the same harness reaches the wall budget in five of
its eighteen cells, so its failures there are neither a reply-contract failure
nor predominantly a speed failure. That rung is the cross-family point, so the
change tracks the family, not the size.

Repetition is the second failure shape. In many cells the agent repeats one
action verbatim until the cell ends: at 0.75B the median cell's longest run of
identical calls is 272 for pi and 36.5 for OpenCode against 21 for \system{}, and
pi's longest run in one cell reaches 273. At 1.9B the medians are 63.5 and 2.5
against 21. \system{} repeats as well, but its transcript records an explicit
duplicate-call intercept and its runs stay in the tens, not in the hundreds.
Pooling the three CLI baselines within a rung, cells whose longest run reaches
ten calls average 0.09, 0.25, 0.01 and 0.55 at the four rungs, against 0.27,
0.44, 0.45 and 0.83 for the cells below that threshold. The relation is a
correlation, not a treatment effect, because the harder tasks produce both
longer runs and lower scores.

The third shape concerns what a score means. The scorer reads structure and
fixture strings, not provenance, so a fluent artifact can pass with invented
citations. On HWF-01, the one task whose fixture supplies external DOIs, we
audited every DOI in the produced artifacts against the eighteen records the
fixture ships. At 0.75B, pi's artifact carries those eighteen DOIs and no
others. \system{}'s carries twenty-eight distinct DOIs, of which six appear in
the fixture, so twenty-two are invented. Both cells score 1.000. The scorer
never reads a DOI, and on this cell its verdict cannot distinguish the two. The
audit therefore bounds what a high score means for a small model without
contradicting any score here. An artifact that names the right files and
sections can still carry four invented identifiers for every one it quotes.

A blind audit of sixty cells by two reviewers puts a number on that depth. Their
mean judgment correlates with the checker at 0.57 ($p<0.001$), and on the
thirty-four cells the checker scores 1.000 both reviewers independently mark the
content wrong in eight. A high score here is conformance to the specification,
not correctness of the work.

\subsection{Cost and the upper anchor}
\label{sec:cost}

Every cell records its wall clock. The five arms share one server that executes
a single request at a time, so the elapsed time of a rung is a fraction of its
total and GPU occupancy is lower still.

The whole 0.75B rung cost 6.1 agent-hours for ninety cells, the 1.9B rung 9.6,
3.7B 19.4 and 4.7B 6.5. Per-cell time and score do not travel together: at
0.75B OpenCode has the shortest mean cell at 2.4 minutes and one of the lowest
scores, and at 4.7B the five arms finish within a minute of each other.

Per-cell minutes hide the cost that decides a deployment, which is the time
spent per artifact set actually delivered. Table~\ref{tab:cost} divides each
arm's total time in a rung by its number of complete artifact sets. At 3.7B
OpenCode needs 195 minutes per complete set against the winner's six, a factor
of thirty-two, although its per-cell mean is only four times longer. At 0.75B it
never produces a set at all, and the arm that spends the least time per cell
there is the one that finishes nothing. The winner is also the cheapest arm per
delivered set at all four rungs, and the others do not keep their score order:
at 1.9B pi is second by score and slowest of the four arms that deliver a set,
while goose is second by cost there and third by score.

\begin{table}[t]
\centering
\caption{Wall-clock minutes of agent time per completed artifact set: an arm's
total time in the rung divided by the number of its cells that wrote every
declared artifact. A dash means the arm never completed a set at that rung.
These are single-tenant numbers on one server handling one request at a time.}
\label{tab:cost}
\small
\begin{tabular}{@{}lccccc@{}}
\toprule
Rung & \system{} & pi & OpenCode & goose & mini-swe-agent \\
\midrule
0.75B & 10 & 32 & -- & 27 & -- \\
1.9B & 11 & 29 & 19 & 15 & -- \\
3.7B & 6 & 10 & 195 & 64 & 70 \\
4.7B & 5 & 6 & 11 & 7 & 72 \\
\bottomrule
\end{tabular}
\end{table}

The 35.5B anchor bounds the measurement at the top instead of extending the
ladder. Both arms were run on all eighteen tasks, 36 cells in total, and the
anchor is reported on the six cheapest, chosen as the ones with the fewest
deliverables. goose solves all six and averages 1.000. \system{} solves five and
averages 0.833, because one cell reaches its budget. Over all eighteen tasks the
two arms average 0.952 and 0.944, and the paired difference on the six tasks is
$-0.167$ with an interval of $[-0.500,0.000]$, which is not separable from zero
at six tasks. The anchor shows the ceiling: at 35.5B one harness solves
the cheapest six outright and the other is one budget short, so the task set has
a defined upper working range. The wall clock says the same thing from the other
side: goose spends 40.0 minutes over the six cells and \system{} 95.8, with the
longest cell 10.0 minutes against 45.0. Those minutes come from the hardware rather than the harness.
Only a quarter of the model's weights fit in the card at this context (5.5 of
22.6\,GB), so most of the inference runs on the CPU and the anchor's wall clock
is not comparable to the ladder's. A subset chosen for runtime is a subset
chosen for ease, and the anchor is reported as a bound on the measurement, not
as evidence that harness differences vanish at the top of the ladder. In the
earlier matrix the 4.7B Qwen outscored this model over the same eighteen tasks
\citep{mingbird2026}, and that ordering does not survive on the six cheapest
tasks.

\subsection{Precision}
\label{sec:precision}

The ladder above changes the model and holds the harness fixed. It does not hold
precision fixed, and Section~\ref{sec:threats} states the consequence: the two
smallest Qwen tags ship at Q8\_0 and the rest at Q4\_K\_M, so a size comparison
across 1.9B also crosses a bit-width change, and the sign of that change cannot
be read from the ladder because a wider model and a better-quantized one move in
the same direction. The interaction is known from the quantization literature,
where bit width and parameter count trade off against each other and the
trade-off is not linear \citep{dettmers2023case, kumar2024precision}.

To measure it we ran a precision ladder on a single base model at four bit
widths: \texttt{granite4.2:3b} at q2\_K (3.15 bits per weight), q3\_K\_S
(3.63), Q4\_K\_M (4.85) and q6\_K (6.50). Bits per weight are computed
from each served artifact's byte count and parameter count, not from the label,
because a K-quant carries more bits than its name suggests. Every level runs the
same 18 tasks with the same two arms and the same protocol as the ladder above:
temperature zero, thinking disabled, a 32K window, and the per-tier budgets of
Section~\ref{sec:design}. The arms are \system{} and pi, the baseline that is
strongest at three of the ladder's four rungs.

The levels are ordered so that every task meets all four bit widths back to
back. Collected level by level, the four would be weeks apart, and a level
collected after a reboot would differ from one collected before it for reasons
that have nothing to do with precision.
Ordering by task makes each of the four comparisons a within-task contrast, and
the arm and level orders both reverse by task parity, so neither position nor
warm-up systematically favors one side. Every level mean reported here therefore
comes from one contiguous, task-interleaved collection.

The hypothesis registered for this axis is that the harness advantage over the
baseline grows as precision falls. The tests are four paired contrasts, one per
level, plus one slope of the contrast against bits per weight. Tasks that both
arms solve at a level are dropped before the contrast, so a ceiling cannot
produce a trend, and where the desaturated and raw readings differ the
desaturated one governs. A level on which every arm scored zero would be a
collapse point and would leave the fit. None of the four reaches that state, so
all four enter it. The family is Holm-corrected separately from
the confirmatory family above, and all five tests stay in it. After correction
only the slope survives, at adjusted $p=0.048$, and the three level contrasts
rise to 0.057. The
lowest level yields only two non-zero pairs, too few for a rank test, so its
contrast carries $p=1$ rather than being dropped: removing a test from a
pre-registered family after seeing the data is the one move the correction
exists to prevent. It stays in the table as well, because a collapse is a
result.

\begin{table}[t]
\centering
\small
\setlength{\tabcolsep}{4pt}
\caption{The precision ladder: one base model at four bit widths, eighteen
tasks, two arms. \emph{bpw} is computed from each served artifact's byte count
and its parameter count, not from the label. \emph{Gap} is \system{} minus pi
and the intervals resample the 18 tasks. The first block is the deployment
reading, in which a cell that reaches its wall budget scores zero. The second
drops tasks both arms solve at that level, which is the reading the slope is
fitted to, and $n$ is the number of tasks left. The third rescores each timed-out
cell from its archived working directory.}
\label{tab:quant}
\footnotesize
\setlength{\tabcolsep}{3.5pt}
\begin{tabular}{@{}lccccr@{}}
\toprule
Level & bpw & \system{} & pi & Gap [95\% CI] & $n$ \\
\midrule
\multicolumn{6}{@{}l}{\emph{Deployment reading}}\\
q2\_K    & 3.15 & 0.029 & 0.000 & $+0.029$ [$+0.000$,$+0.075$] \\
q3\_K\_S & 3.63 & 0.458 & 0.195 & $+0.264$ [$+0.094$,$+0.437$] \\
Q4\_K\_M & 4.85 & 0.822 & 0.493 & $+0.329$ [$+0.131$,$+0.544$] \\
q6\_K    & 6.50 & 0.860 & 0.568 & $+0.292$ [$+0.099$,$+0.503$] \\
\midrule
\multicolumn{6}{@{}l}{\emph{Deployment reading, tasks both arms solve dropped}}\\
q2\_K    & 3.15 & 0.029 & 0.000 & $+0.029$ [$+0.000$,$+0.075$] & 18 \\
q3\_K\_S & 3.63 & 0.458 & 0.195 & $+0.264$ [$+0.098$,$+0.435$] & 18 \\
Q4\_K\_M & 4.85 & 0.708 & 0.171 & $+0.538$ [$+0.261$,$+0.803$] & 11 \\
q6\_K    & 6.50 & 0.747 & 0.221 & $+0.526$ [$+0.225$,$+0.821$] & 10 \\
\midrule
\multicolumn{6}{@{}l}{\emph{Rescored reading}}\\
q2\_K    & 3.15 & 0.029 & 0.029 & $+0.000$ [$+0.000$,$+0.000$] \\
q3\_K\_S & 3.63 & 0.458 & 0.195 & $+0.264$ [$+0.100$,$+0.439$] \\
Q4\_K\_M & 4.85 & 0.877 & 0.854 & $+0.023$ [$-0.167$,$+0.203$] \\
q6\_K    & 6.50 & 0.860 & 0.855 & $+0.004$ [$-0.162$,$+0.180$] \\
\bottomrule
\end{tabular}
\end{table}

Table~\ref{tab:quant} reports the four levels. Under the deployment reading,
which is the paper's primary one, the advantage does not grow as precision
falls. It is largest at 4.85 bits, where \system{} averages 0.822 against pi's
0.493, a gap of $+0.329$ with an interval of $[+0.131,+0.544]$. At 6.50 bits it
is $+0.292$ $[+0.099,+0.503]$, and at 3.63 bits $+0.264$ $[+0.094,+0.437]$. Both
intervals exclude zero. At 3.15 bits \system{} averages 0.029 from fourteen
completed cells and pi's mean is zero because every one of its cells reached its
budget. The gap is $+0.029$ $[+0.000,+0.075]$.

\begin{figure}[t]
\centering
\includegraphics[width=\linewidth]{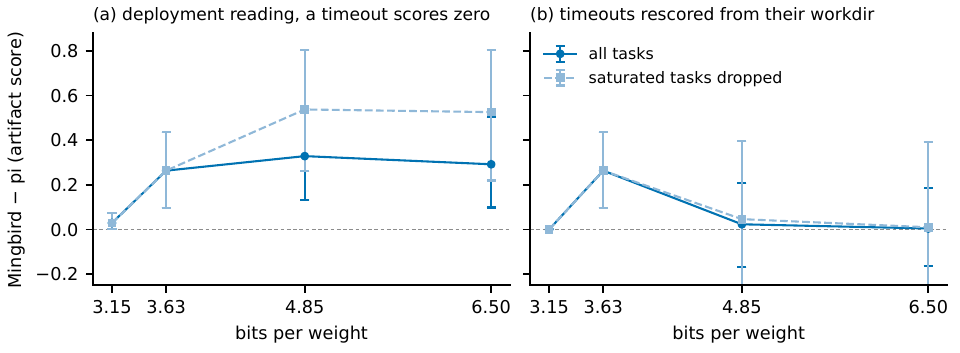}
\caption{The precision ladder under its two readings. Markers are level means
of the paired gap and bars are 95\% task-bootstrap intervals. The filled series
uses all eighteen tasks. The lighter dashed series drops the tasks both arms
solve at that level, which is the contrast the slope is fitted to, so the
second series is the one the test reads. Under the deployment reading (a) the
gap is largest at 4.85 bits and collapses as precision falls, because
compression costs both arms the ability to finish. Under the rescored reading
(b) the two high-precision levels move toward zero, because the cells pi lost
there had produced work that the rescoring credits.}
\label{fig:precision}
\end{figure}

The registered hypothesis, that the harness advantage widens as precision
falls, is contradicted under the deployment reading and undecided under the
rescored one. The gap between the two readings is itself the result: at high
precision the baseline's deficit is time rather than quality.

Fitted across the four levels on the desaturated contrasts, the slope of the gap
against bits per weight is $+0.139$ with an interval of $[+0.037,+0.235]$ and a
bootstrap $p=0.010$. The preregistered direction was negative, a gap that widens
as precision falls. The estimate is positive and its interval excludes zero, so
under the deployment reading the registered hypothesis is contradicted rather
than undecided. Under the Holm correction over all five tests the slope is the
one member of the family to survive (adjusted $p=0.048$). The three level
contrasts follow at $0.057$ each and the untestable level at 1, so the levels
are not significant on their own. The sign is not carried by the collapse point:
dropping the 3.15-bit level makes the estimate more positive.

The rescored block of Table~\ref{tab:quant} takes each of the forty-one cells
that reached its budget and scores the working directory it left behind. It
moves the two high-precision levels and little else. At 4.85 bits pi's mean
rises from 0.493 to 0.854 and the gap falls from $+0.329$ to $+0.023$, interval
$[-0.167,+0.203]$. At 6.50 bits pi rises from 0.568 to 0.855 and the gap falls
from $+0.292$ to $+0.004$, interval $[-0.162,+0.180]$. At 3.63 bits pi's mean is
unchanged and the gap stays $+0.264$ $[+0.100,+0.439]$. At 3.15 bits pi's
rescored mean rises to 0.029 and the gap is zero to three decimals.

So the contradiction of the registered precision hypothesis belongs to the deployment reading, and the
section does not read it as a statement about the systems alone. Fitted to the
rescored gaps the slope is $-0.030$ with an interval of $[-0.152,+0.098]$, which
is indistinguishable from zero: if the work already on disk when the run was
stopped counts, this design cannot order the levels at all. What holds under both
readings is smaller and still useful. At 3.15 bits both arms collapse to
near-zero scores: \system{} keeps a mean of 0.029 from the fourteen cells it
finishes, while every pi cell reaches its budget. At 3.63 bits both produce work
but at a fraction of their higher-precision scores, 0.458 and 0.195. Between 4.85 and 6.50 bits pi's deficit is a deficit of time and
not of quality, since its timed-out cells were completing most of the task. That
is the signature the rescored reading is built to measure.

The arms also fail differently at every level. At 6.50 bits pi reaches its
budget in seven of eighteen cells and \system{} in none. At 4.85 bits eight of
pi's eighteen reach it, against one \system{} timeout and one cell that exits on
an error in the harness. At 3.63 bits six of pi's cells reach it and one stalls,
against none for \system{}. At 3.15 bits all eighteen of pi's cells reach it
while \system{} finishes fourteen and produces no artifact in three. A timeout scores
zero under the deployment reading, so part of the low-end gap measures whether
work finishes at all rather than how well it is done. Cells that reached the budget stay in the
denominator, scored zero, which is the deployment rule used everywhere else in
the paper.

The four levels are four quantizations of one checkpoint, all official GGUF
files downloaded from the publisher rather than re-quantized here, so no level
is a second compression of another. The base model is granite 4.2 3B, the only
family in the study served with a full range of bit widths: the 4.7B family
ships at two levels, which cannot carry a four-level ladder. The top level is
Q6\_K rather than Q8\_0 because at a 32K window Q8\_0 holds 6.28 of its
6.76\,GB in VRAM and would have measured memory placement alongside bit width.
The comparator is pi, the stronger baseline at 3.7B (0.714 against goose's
0.262), so the contrast is if anything harder than one against goose. The
ladder is therefore a within-family measurement: four quantizations of one
checkpoint at one rung, and its result does not transfer to another rung by
construction. It ran on the successor host of \ref{app:repro}.

Read as a measurement of the caveat rather than as a trend, the pattern is that
compression costs the two arms different things. Going from 6.50 to 4.85 bits
costs pi 0.075 under the deployment reading and \system{} 0.038. At 3.63 bits both fall to a fraction of their
high-precision scores and at 3.15 bits both reach zero. The ladder does not support the
claim that the margin is larger where deployments actually run:
under the deployment reading it is, and under the rescored reading it is not,
because the baseline's losses at those levels are work already on disk when the
budget ran out rather than work that never existed. A reader on a fixed overnight schedule
should read the harness choice at high precision as a choice about score. A
reader who can wait for late work should read it as a choice about time.

\section{Discussion}
\label{sec:discussion}

\paragraph{Harness choice sets the level at every size} The result a
practitioner can act on is the level, and it is the same result at every rung.
\system{} is first at 0.75B, 1.9B, 3.7B and 4.7B. Of the four pre-registered
rung contrasts two survive Holm correction, at 0.75B (adjusted $p=0.0067$) and
4.7B ($0.0410$), and two do not. Thirteen of the sixteen paired comparisons
exclude zero overall, and the leads that do run from 0.075 over the strongest
rival at 4.7B to 0.819 over the weakest. A single rung's level is not a stable
object: the 0.75B winner averages 0.664, 0.503 and 0.431 in three collections,
and the same arm at granite averages 0.762 in the matrix and 0.822 in the
precision ladder. The ordering is steadier
than the level, but only for the pair that was repeated: Mingbird and goose were
re-collected at 0.75B and 4.7B and Mingbird leads both times, while pi, OpenCode
and mini-swe-agent were run once and their levels rest on a single collection. A
team that owns one 8\,GB card should choose the harness with as much care as the
model, and should not expect a larger model to repair a bad one: at 4.7B the
four baselines still sit 0.075 to 0.819 below the winner.

\paragraph{The slope is not measurable with four rungs} Two published predictions
disagree about the direction of the interaction, one expecting the strongest
model to gain most from scaffolding and one expecting the gains to shrink as
capability rises. This ladder cannot settle the disagreement, and it shows why
the disagreement is hard to settle at all. Three artifacts each produce a
direction. A floor arm that fails at every size inflates the span into a rise. A
ceiling at 4.7B, where \system{} and pi both score 1.000 on eleven of the
eighteen tasks, compresses the winner's
margin into a decline. A single cross-family rung at 3.7B puts a
peak in the middle. Each of the three has been read as a property of harness
leverage, and each is reproducible here by choosing the estimator. A design that
could resolve the slope needs tasks that stay unsaturated at the top and a
baseline set with no floor arm. That is a different suite, not a longer run of
this one.

\paragraph{What the ablation rules out} The paper's design hypothesis was that the
six executable checks would carry the advantage. The ablation says otherwise at
three of the four rungs. All
six together are worth 0.080 at 0.75B and 0.117 at 1.9B, with intervals that
include zero, and nothing at 3.7B and 4.7B. At the smallest rung the point
estimate is a sixth of the 0.508 lead. The gate
does do its job where the model stops leaving work: at 1.9B, switching it off
turns 0 empty runs into 3. At 0.75B it changes neither the number of empty runs
nor much of the score, so the 0.080 appears in partial credit rather than in
saved runs.

The lead is a progress effect. At 0.75B
between half and three quarters of each gap is cells the baseline left empty,
and most of the rest is a depth difference on tasks both arms completed. At 1.9B
almost all of it is availability. Against pi at 4.7B, where no cell is empty,
all of it is depth on tasks both arms finish. The transcripts show the process
behind the availability part: at 0.75B nine of pi's empty cells end in one
failing shell command repeated until the 32K window fills. We name this the
repeat-until-window-full failure. The same model completes the task in
\system{} after the repeat is interrupted and the context reset.
At 0.75B the dominant failure is not wrong work but unbounded repetition of
one failing action. A harness that bounds repetition turns empty runs into
completed ones, and the completion gate is orthogonal to that channel. The
machinery that bounds it is what the completion gradient does not touch: a tool
surface that takes the directory as a call rather than a shell string, a prompt that states the platform and the
paths, loop and context guards, and compaction that keeps the window usable. One
piece of it can be switched off alone, and the test is not subtle: without the
streak rule the score falls by 0.411 at 0.75B and eight of eighteen tasks lose
their artifact altogether. Those are the parts to replicate. A
completion gate is cheap and harmless, and it is the right response to a model
that cannot recognize its own incomplete work, but a study that measured only
the gate would report a small effect and could conclude, wrongly, that harnesses
do not matter.

\paragraph{The failure to design against is the quiet stop} Exit status, the
transcript's error line and the artifact state are three signals that do not
substitute for one another, and a failure gets harder to see the further back
the scorer looks. A benchmark that
scores crashes will rate an agent highly while it produces nothing, because the
typical failure of a small model is a normal exit with an empty working
directory. Of the 127 empty cells, 85 end normally and 40 of those carry no
error line at all. This changes what an evaluation should measure. Artifact
state at the end of a run is the outcome of interest, and the process exit is
not. It also changes what a harness should do. A stop rule that trusts the
model's claim of completion is a stop rule optimized for a model that is usually
right, which is not the model in front of it.

The cost of getting this wrong is asymmetric. A harness that stops too early
loses the whole task. A harness that stops too late loses time, and time is what
the budget measures. Completion control buys the first against the second, and
the price is visible in the cost numbers: at 1.9B \system{} spends 4.2 minutes
per cell more than OpenCode and earns nine more complete artifact sets.

\paragraph{What the upper anchor does and does not show} At 35.5B goose
saturates the six-task subset and \system{} is one budget short of it, so at that
size the suite has no resolution left for one arm while the other stops just
short of the ceiling. That is a property of the measurement, not a finding
about large models. A ceiling anchor that covers only the cheapest tasks cannot
say whether harness differences persist at scale. It does say that the
task set has a defined upper working range, which is
useful information for anyone reusing it.

\paragraph{Relation to the Mingbird system paper} The preprint
\citep{mingbird2026} describes \system{} as an architecture: its components, its
seven design principles, its completion-control mechanism, and a 288-cell
development matrix over four models from 2B to 35B. This paper treats the same harness as one of
five arms and asks a different question, which is how the choice of harness
compares across model sizes. How \system{} is built is described there, and what
it is worth against four alternatives at four sizes is measured here.

\section{Threats to validity}
\label{sec:threats}
\paragraph{Construct validity} The outcome is artifact completion on 18 tasks
scored by a deterministic checker. That is a narrow slice of what practitioners
want from an agent. It rewards a working directory that satisfies a written
specification and says nothing about code review, maintainability, or the
quality of the reasoning that produced it. The prompt states which files are
expected and which checks will run, so the tasks measure specification
following. The checks are shallow by construction, since they test for named
files, required sections, and values quoted from the fixtures. A hollow artifact
that names the right things can pass, and Section~\ref{sec:failure-results}
measures this on \mbox{HWF-01}, where 22 of the 28 distinct DOIs in one artifact were
invented while the task still scored 1.000. A deterministic checker is a partial
oracle in the sense the software-testing literature gives that term
\citep{barr2015oracle}: it decides the cases it can decide and is silent about
the rest, and a study built on one inherits exactly that limit. The instrument
properties reported in Section~\ref{sec:design} are the ones the current
checklist for agentic benchmarks asks for \citep{zhu2025abc}. The checker was
audited against blind human scores. Sixty cells were drawn at random from this
collection, from the 202 whose declared artifacts were still on disk to show
reviewers, 230 having produced at least one at collection time, and two
reviewers scored each for completeness, correctness and use of evidence. Their
quadratic-weighted agreement runs from $\kappa=0.58$ to $0.69$, and their mean
correlates with the checker at 0.57 (permutation $p<0.001$). What the
correlation does not repair is the depth of the check. In 8 of the 34 cells that
the checker scores 1.000, both reviewers independently mark the content wrong,
and across those 34 cells their mean sits at 0.56 of the scale. The score
indexes conformance to a written specification rather than perceived quality.
The human scores return the same five arms in the same ranking except that goose
and \system{} trade the top two places, and every pair of their bootstrap
intervals overlaps. The suite is also ours, and
the gaps reported here cannot be
separated from authorship. What would separate suite fit from a genuine harness
advantage is a task family written by someone with no stake in any of the five
harnesses. No independent task family was available, though the released suite allows
one to be added. Every task, fixture and scorer is released so that
such a family can be placed beside this one. A ceiling follows from the same
shallow checks: at 4.7B \system{} and goose both reach 1.000 on eight of the
eighteen tasks, so more than a third of the suite cannot separate them there.

\paragraph{Internal validity} Each harness ships its own prompt and tool surface,
so the comparison is between products and a difference between two of them
cannot be attributed to one design decision. The mechanism gradient is the only
place where a single family of mechanisms is switched with everything else
fixed. Each cell is a single run. The gradient's full arm reruns the matrix's
\system{} cell at the same rung: the two differ by 0.233, 0.229, 0.148 and 0.012
at 0.75B, 1.9B, 3.7B and 4.7B, and a second collection of the whole 3.7B
gradient moves that rung by 0.066. An arm mean can therefore move by nearly a
quarter between collections with the configuration unchanged, so a point difference
smaller than that is not interpreted across collections, and every paired
contrast is computed inside one collection. A rerun of 72 individual cells
disagreed with the matrix on 29 of 67 comparable cells, but the median absolute difference
is zero and the disagreement is bimodal: a task usually repeats its score
exactly and occasionally lands a whole point away. Per-cell noise is carried by
the paired task-level bootstrap, which widens each interval accordingly. The rungs were
collected in sequence, so a size trend is confounded with calendar time. Within
a rung the harness order alternates by task, which balances warm-up and drift
across harnesses but not across rungs. Quantization is not constant: the two
smallest Qwen tags ship at Q8\_0 and the rest at Q4\_K\_M, so the change falls
between 1.9B and 3.7B. Its direction is measured in Section~\ref{sec:precision}
on one base model at four bit widths, and it is not a constant offset: the
baseline loses more than the completion-controlling harness as precision falls
from 6.50 to 4.85 bits. At 3.63 bits both produce at a fraction of their
high-precision scores, and at 3.15 bits both collapse to near-zero scores, with
pi reaching its budget in all eighteen cells while \system{} finishes fourteen
and produces nothing in three. A bit-width change across 1.9B therefore cannot
be subtracted from a size comparison as a fixed quantity.

A cell that reaches its budget is scored zero under the deployment reading, so a
slow but correct harness is penalized. The rescored reading replaces those zeros
with the score of the archived working directory, and both readings are reported
at every contrast. The rescored reading conditions on an outcome of the harness,
because whether a cell produces anything is part of what the harness determines,
so it describes where the scores come from and is not used to rank arms on its
own.

One arm's floor is a reply contract and not a graded capability gap.
mini-swe-agent ends fifteen of eighteen cells at 0.75B and fifteen at 1.9B with
its own \texttt{Repeated\allowbreak FormatError}, and at 3.7B five of eighteen
reach the wall budget instead.
The five-arm span therefore measures what a harness whose contract a model
cannot meet costs to deploy, as well as graded differences between harnesses
that can meet theirs. The span without that arm is reported beside the raw span
at every rung.

\paragraph{Cell exclusions} A ladder cell is excluded only when the harness
binary failed to start, when the runner raised an exception before the agent's
first model call, or when the inference server returned a 5xx error during the
cell. These are infrastructure failures and not agent outcomes, and the affected
cell is re-run. Three of the 360 ladder cells meet the rule. The excluded cells
are listed with their reasons in the analysis output, and the reported means are
also given with the exclusions reversed there; the reversal moves the affected
rung means by 0.013 and 0.023 and leaves every conclusion unchanged. Cells in
which the agent ran and then crashed are not excluded. They are scored as
failures.

\paragraph{External validity} The ladder and the gradient ran on one Windows 10
machine (an Intel Xeon E5-2667 v2, 48\,GB of RAM, an NVIDIA RTX 5060 with
8\,GB, and one Ollama 0.40.6 server), so the hardware is a workstation with a
consumer GPU and not a laptop. All four rungs fit entirely in the 8\,GB card at the study context. Four
models cover the band where local deployment happens, but they are not a random
sample of open-weight models, and the results need not transfer to larger
models, other runtimes, or other task suites.

One asymmetry shapes the failure mix. The host is Windows, the workspace path
contains a space, and the third-party harnesses invoke a Unix shell. Windows is
a distinct agent environment with its own benchmark \citep{bonatti2024windows},
and the mismatch here belongs to the environment rather than to the harnesses. A
model that has learned Unix commands writes an unquoted path, and the shell answers
``cd: too many arguments'': pi records that error in eleven of its eighteen
cells at 0.75B, OpenCode in fifteen across the ladder, six at 0.75B and six at
3.7B, while neither
\system{} nor goose records it at any rung. \system{} is the only arm that states
the platform and the shell mapping in its prompt, and it is not immune to the
mismatch: at 0.75B five of its eighteen transcripts mention a Unix-rooted path.
The asymmetry is part of what the system is, so we do not remove it, and the
small-end results should not be read as platform-independent. It is, however,
measurable against the contrasts. Dropping every task on which the baseline
never writes the unquoted path leaves the 0.75B gap to OpenCode at $+0.494$
$[+0.228,+0.740]$ over eleven tasks against $+0.546$ over seventeen, and the
3.7B gap to OpenCode at $+0.763$ $[+0.525,+0.954]$ over twelve against $+0.762$
over seventeen. Against pi the exclusion leaves $+0.367$ $[-0.041,+0.714]$ over
the seven tasks pi never fails this way, against $+0.483$ over seventeen. Seven
tasks are too few to settle that pair. The mismatch is therefore part of what
separates the arms at 0.75B and is not demonstrably its carrier: the gap to
OpenCode survives the exclusion and the gap to pi is left unresolved rather than
overturned.

We did not run a second benchmark. The benchmarks that are standard at this scale
($\tau^2$-bench, GAIA) need a cloud user simulator or a judge model, which would
add a second model and a second source of variance to a study whose point is to
hold everything except the harness fixed.

\paragraph{Conclusion validity} Eighteen tasks per cell bound the precision of
every contrast, and a null is read as smaller than the study could see. At 4.7B
the observed contrast with goose, $+0.250$, sits just above that rung's minimum
detectable effect of 0.221 and is significant all the same, which is not a
contradiction: the minimum detectable effect is the difference a study of this
size detects four times in five, not a floor below which nothing can be found.
The confirmatory family has seven tests and is corrected with Holm. The seven
are not independent, since the four per-rung contrasts share tasks and arms and
the three trend tests are computed from the same per-rung quantities. Holm holds
under arbitrary dependence, so the sharing does not invalidate the correction.
It makes the corrected values conservative when the tests move together, which
they do here.

\paragraph{Experimenter bias} The authors wrote one of the five harnesses, the 18
tasks, the scorer and the analysis plan. Four things limit the room for that
bias. The scorer reads artifacts and not the transcript. Every cell is
recorded. Budgets and sampling are identical
within a rung. The analysis plan
and its multiplicity family were fixed before the confirmatory contrasts were
read, as Section~\ref{sec:analysis} records. The failure coding is mechanical, a
file-existence test and the runner's termination record, but it was not coded
blind to harness identity because the directory names carry it. Every number
can be recomputed from the per-cell results, and any harness can be rerun
against the same suite.

\section{Conclusion}
\label{sec:conclusion}

Local agents on consumer hardware are usually discussed as a model problem. They
are also a harness problem. This study compared five harnesses on 18
artifact-producing tasks against four open-weight models from 0.75B to 4.7B on
one Windows host, with a 35.5B anchor at the top.

The completion-controlling harness had the highest mean at each of the four
rungs. Two of the four pre-registered rung contrasts survive Holm correction and
thirteen of sixteen paired comparisons exclude zero, with the leads that do run
from 0.075 to 0.819. That ordering is what repeats. Three collections of the
winner's 0.75B cells average 0.664, 0.503 and 0.431, and in the two that also
contain a baseline the paired gap against goose is $+0.508$ and $+0.476$.

The slope of that lead is a different question, and four rungs cannot answer it.
The span slope is $+0.198$ $[-0.078,+0.476]$, consistent with a fall of a tenth
of a point per decade and with a rise of almost half a point. The three
artifacts in Section~\ref{sec:shape} show why any answer would be an artifact of
the estimator: a floor arm produces a rise, a ceiling a decline, a cross-family rung a peak.

The ablation contradicts the design story. The six executable
completion checks, which are what the harness was built around, are worth 0.080
at 0.75B and 0.117 at 1.9B, with intervals that include zero, and nothing at the
two large rungs. At the smallest rung the point estimate is a sixth of the lead.
At 1.9B it is half, and there the checks are what turn three empty runs into
finished ones. At the largest the interval bounds their share at 18\%. The rest
is a progress effect: the tool surface sends a directory listing through a shell that
rejects the unquoted path, and the model then repeats the failing call until the
context window fills. At 0.75B pi leaves nine empty cells and every one of them
carries that shell error, and the same model completes the same task in \system{} once
the repeat is interrupted and the context reset. That rule is tested directly:
removing it and nothing else costs 0.411 at 0.75B and turns eight of eighteen
tasks from an artifact into an empty directory. The design still does not
separate the tool surface from the rule, since the failing shell call that
starts the loop is a consequence of the surface, and a tool-surface ablation
inside one harness remains the next study.

Of 127 cells that produced none of their declared artifacts, 85 ended
with a normal exit and 40 of those recorded no error line. A benchmark that
read the exit status as the outcome would credit every one of them, and the fix
is to score the artifact, not the process.

Three things would strengthen the design: a task suite written by someone with
no stake in any of the five harnesses, a second host operating system, and
repeated runs per cell. A fourth was measured here rather than assumed.
Quantization is not a constant across the ladder, so Section~\ref{sec:precision}
runs one base model at four bit widths and reports where the advantage survives
compression and where it does not. Every cell, transcript, and score is
available from the corresponding authors, so that a reader can attempt the
three extensions this study could not run.

\section*{Declarations}

\paragraph{CRediT authorship contribution statement}
{\hyphenpenalty=10000\exhyphenpenalty=10000
These two authors contributed equally to this work and are co-corresponding
authors. Hao Wang: conceptualization, methodology, software, investigation,
formal analysis, writing -- original draft. Ting Huang: conceptualization,
methodology, validation, writing -- review and editing, supervision.\par}

\paragraph{Declaration of competing interest}
The authors declare the following financial interests or personal relationships
which may be considered potential competing interests. Ting Huang is an employee
of Honor Device Co., Ltd. The authors wrote one of the five harnesses evaluated
in this study, and Section~\ref{sec:threats} discusses how far this could affect
the results. The other four harnesses are third-party software used under
their own licenses, and the company had no role in the design, execution, or
reporting of the study.

\paragraph{Funding}
This research received no external funding.

\paragraph{Ethics approval}
Not applicable. The study involves no human participants, human data, or
animals.

\paragraph{Data availability}
The held-out task suite and fixtures, the deterministic scorer, and the runner
are public at \url{https://github.com/Mingbird/Mingbird-agent}. Every number in
this paper is backed by one score file, transcript and working directory per
cell, together with the analysis scripts and the digests of the served model
artifacts. That archive is too large to carry as supplementary data and is
available from the corresponding authors on request.

\paragraph{Declaration of generative AI in writing}
Large language models were used to assist with code development, experiment
orchestration, statistical analysis, and language editing. All technical claims,
numbers, and interpretations were reviewed and verified by the authors, who take
full responsibility for the work.

\appendix
\section{The held-out task suite}
\label{app:tasks}

The suite has eighteen tasks in four tiers. Each task is a working directory
with fixtures, a prompt, a list of milestone checks, and a list of final
deliverables. The prompt states the deliverable names and the checks that will be
applied, and the fixtures contain the data the task is about. Nothing in the
working directory tells the agent which harness it is running under.

\newcommand{\tabbodyE}{%
\footnotesize
\begin{tabular}{@{}llllp{6.2cm}r@{}}
\toprule
Task & Tier & Work type & Fixtures & Principal deliverables & Checks \\
\midrule
HWF-01 & Retrieval & materials recent literature map & 1 & \mbox{recent\_review.md}, \mbox{citation\_index.md} & 4 \\
HWF-02 & Retrieval & manuscript integrity audit & 1 & \mbox{revised\_conclusions.md}, \mbox{verification\_checklist.md} & 4 \\
HWF-03 & Synthesis & cli feature per day report & 1 & \mbox{timetrack.py}, \mbox{usage\_day.md} & 4 \\
HWF-04 & Synthesis & characterization first debugging & 2 & \mbox{inventory.py}, \mbox{regression\_notes.md} & 4 \\
HWF-05 & Synthesis & quarantine cleaning pipeline & 1 & \mbox{tidy\_data.csv}, \mbox{rejected.csv}, \mbox{sensor\_rollup.csv}, \ldots & 6 \\
HWF-06 & Synthesis & hardness microstructure analysis & 1 & \mbox{fig\_hardness.png}, \mbox{fig\_ductility.png}, \mbox{plot\_hardness.py}, \ldots & 7 \\
HWF-07 & Workflow & web research knowledge base selection & 0 & \mbox{matrix.md}, \mbox{verdict.md}, \mbox{links.md} & 5 \\
HWF-08 & Workflow & api query half design and mock & 1 & \mbox{app\_query.py}, \mbox{demo.md} & 4 \\
HWF-09 & Workflow & benchmarked refactor & 1 & \mbox{fast\_impl.py}, \mbox{bench\_report.md} & 5 \\
HWF-10 & Workflow & cli cheatsheet documentation & 1 & \mbox{usage\_examples.md}, \mbox{failure\_modes.md}, \mbox{cli\_cheatsheet.md} & 5 \\
HWF-11 & Workflow & tco driven product selection & 1 & \mbox{decision\_memo.md}, \mbox{migration\_risk.md} & 4 \\
HWF-12 & Workflow & research software course design & 0 & \mbox{sessions/session01.md}, \mbox{assignments/project\_brief.md}, \mbox{readings.md} & 5 \\
HWF-13 & Workflow & privacy by design refactor & 1 & \mbox{payments\_v2.py}, \mbox{controls\_checklist.md} & 4 \\
HWF-14 & Workflow & qa and rollout planning & 1 & \mbox{rollout\_plan.md}, \mbox{risk\_register.md} & 5 \\
HWF-15 & Workflow & alternate method reproducibility audit & 2 & \mbox{ea\_recompute.py}, \mbox{fig\_offset.png}, \mbox{audit2\_report.md} & 5 \\
HLH-01 & Long horizon & long horizon mean aggregation chain & 1 & \mbox{means\_aggregate.py}, \mbox{means\_check.json}, \mbox{synthesis.md}, \ldots & 14 \\
HLH-02 & Long horizon & long horizon context pressure variant & 1 & \mbox{stream\_stats.py}, \mbox{wordlen\_index.json}, \mbox{topic\_counts.md}, \ldots & 7 \\
HLH-03 & Long horizon & long horizon endurance reverse & 11 & \mbox{m01.py}, \mbox{m02.py}, \mbox{m03.py}, \ldots & 25 \\
\bottomrule
\end{tabular}}
\newlength{\fitwE}
\settowidth{\fitwE}{\tabbodyE}
\begin{table}[t]
\centering
\caption{The 18 held-out tasks. Checks are the milestone and final-artifact checks the scorer runs, and every check reads a file in the working directory.}
\label{tab:btasks}
\resizebox{\linewidth}{!}{\begin{minipage}{\fitwE}\tabbodyE\end{minipage}}
\end{table}

Each task starts from a workflow that a practitioner described in public:
screening a literature export, auditing a manuscript's claims against its own
tables, turning a CLI's logs into a daily usage report, reproducing a published
figure from the raw data, and so on. The task was then written so that completion
is a property of the working directory: every check reads a named file and
applies a rule that can be evaluated without a model. No task was written by
looking at how any harness performed. The suite was frozen before the ladder ran
and no task was edited afterwards.

Tier~1 asks for extraction from a fixture database. Tier~2 asks for synthesis
across records, including disagreements between them. Tier~3 asks for a
multi-step workflow with code, data products and a written deliverable. Tier~4 is
long-horizon, with an intermediate deliverable that later steps consume and, in
two of its three tasks, eight or more milestones. The tiers vary the length of
the dependency chain rather than the domain difficulty, which is why the ladder
is analyzed per task and not per tier.

Every check is binary. The milestone score is the mean of its checks, the
final-artifact score is the mean of its checks, and the task score is
\[
\text{task} = \tfrac{4}{7}\,\text{milestone} + \tfrac{3}{7}\,\text{final}.
\]
The scorer also defines a third, judge-based dimension at weight 0.3. It is
disabled in every reported run, which is why the two deterministic dimensions
carry weights 4/7 and 3/7 rather than 0.4 and 0.3. A cell's outcome is its task
score, and a cell that reaches its wall budget scores zero regardless of what its
working directory contains. HWF-01, for example, has two milestone checks and two
final checks, so a cell that passes one milestone check and both final checks
scores $\tfrac47(0.5)+\tfrac37(1.0)=0.714$.

\begin{table}[H]
\centering
\small
\caption{The 117 checks in the held-out suite by type. The type names are the
scorer's own. A check that only requires a file to exist cannot distinguish a
correct artifact from a plausible one, which is why the paper reports the
identifier-fidelity audit alongside the score. Shares may not sum to 100\%
because of rounding.}
\label{tab:bchecks}
\begin{tabular}{@{}llrr@{}}
\toprule
Check type & What it tests & Count & Share \\
\midrule
\texttt{exists} & the file is present & 44 & 38\% \\
\texttt{contains\_any} & one of a list of strings appears & 25 & 21\% \\
\texttt{python\_compiles} & the file parses as Python & 18 & 15\% \\
\texttt{min\_words} & a word-count floor & 13 & 11\% \\
\texttt{contains\_groups} & several groups of strings each appear & 5 & 4\% \\
\texttt{structure} & required sections and a word floor & 3 & 3\% \\
\texttt{image\_valid} & the image opens and is non-trivial & 3 & 3\% \\
\texttt{min\_urls} & at least $n$ URLs & 3 & 3\% \\
\texttt{not\_contains} & a forbidden string is absent & 1 & 1\% \\
\texttt{spec\_pytest} & the task's own test file passes & 1 & 1\% \\
\texttt{script\_pass} & the script runs and exits zero & 1 & 1\% \\
\bottomrule
\end{tabular}
\end{table}

More than half of the checks test presence or a substring
(Table~\ref{tab:bchecks}), so a score is best read as ``the harness produced the
specified files with plausible content'' rather than ``the work is correct''. The
identifier-fidelity audit in Section~\ref{sec:failure-results} measures the part
the checks cannot.

The tasks do not test conversational ability, tool selection under ambiguity, or
recovery from a hostile user. They test whether an agent can turn fixtures into a
specified set of files. That is a narrow construct, chosen deliberately so that
the outcome can be checked deterministically at every cell.

\section{Reproducibility}
\label{app:repro}

The five harnesses (four third-party at the versions in
Table~\ref{tab:bharnesses}, with \system{} at release v3.1.9, commit
\texttt{56cd23d}), the 18
tasks with their fixtures (CC BY 4.0), the deterministic scorer and the runner
that executes a cell are at \url{https://github.com/Mingbird/Mingbird-agent},
which also holds the earlier 288-cell matrix and the design-paper release. The
scorer and the runner are code and carry the harness's Apache-2.0 license, not
the data license. The per-cell archive, one score file, transcript and working
directory per cell, together with the sampling proxy and the analysis scripts,
is described under Data availability.

\paragraph{Hardware and backend} The matrix ran on one Windows 10 machine: an
Intel Xeon E5-2667 v2, 48\,GB of RAM, and an NVIDIA RTX 5060 with 8\,GB, served
by Ollama 0.40.6 started as a dedicated process on port 11437 with
\texttt{OLLAMA\_CONTEXT\_LENGTH=32768} and \texttt{OLLAMA\_KEEP\_ALIVE=-1}. A
sampling proxy on port 11438 rewrites the temperature and the thinking field of
every request from the harnesses that expose no such knob. The proxy logs each
request, and that log is in the per-cell archive. That thinking is off is verified per rung
rather than assumed: each probe sends one prompt through both proxy paths and once
to the model with thinking forced on, and on every model the study loaded the two
proxy paths return an answer of identical length while the forced-on control
returns reasoning text and no answer. The records are in the per-cell archive as
\texttt{analysis/thinking\_probe\_*.json}.

The loaded model's size, processor split and context are recorded from the
server's \texttt{/api/ps} together with the pid that owned the port, at the end of
every cell and, from 25 October on, at the start as well. The end-of-cell record
reaches every reported cell except the first three.

The served window is verified per cell rather than assumed. The guard records
the pid of the server it starts and evicts any other process holding the port,
\system{} carries its own context through \texttt{AGENT\_CTX}, and the matrix
driver refuses to start a cell against a runner whose window is off the
protocol. An extraction of the server's startup log is in the per-cell archive
as \texttt{engine\_windows.json}, one record per server start with the context
pin it was launched under.

\paragraph{The precision ladder ran on a second host} The matrix, the gradient
and the anchor ran on the host described above. The precision ladder of
Section~\ref{sec:precision} ran on a second machine of the same class: an Intel
Xeon E5-2696 v3 with 64\,GB of RAM, the same RTX 5060 with 8\,GB, and the same
pinned Ollama build and protocol. Every (task, arm, level) triple inside a task
ran on one machine in one contiguous window, so a platform difference cannot be
recorded as an effect of bit width.

\paragraph{One-cell reproduction}
\sloppy
Prepare a clean workdir from the task JSON and
its fixtures, then run \texttt{run\_cell.py} with the agent, task, model, \mbox{timeout},
run id, the two ports and the harness root as arguments. The runner writes
\texttt{score.json}, \texttt{transcript.txt} and the working directory per cell,
so a failed reproduction is inspectable at the same granularity as the published
data. Rescoring a working directory with the scorer that ships at
\texttt{repo/bench/lrab/scoring/score\_task.py}, passing the task, the working
directory and \texttt{-{}-no-judge}, reproduces the published score exactly: every
cell that carries both a score and a working directory was rescored this way and
all of them matched, and the same procedure produced the rescored reading of
Section~\ref{sec:contrasts} for the 58 timed-out cells. Three arms need a binary
we cannot redistribute. pi is a Node package
(\texttt{@earendil-works/pi-coding-agent} 1.3.0) that the runner resolves through
\texttt{PI\_CLI}, then \texttt{\textasciitilde/myagents-bin/pi}, then
\texttt{PATH}. mini-swe-agent runs under a Python interpreter selected by
\texttt{AGENT\_MINI\_PY}, and the shims in \texttt{experiments/artifacts/bin}
point \texttt{opencode} at the 1.18.23 build.

\paragraph{Analysis scripts} \texttt{analyze\_paper\_b.py} computes means, gaps,
trend tests and the minimum detectable effect.
\texttt{analyze\_paper\_b\_gradient.py} computes the mechanism gradient.
\texttt{classify\_failures\_paper\_b.py} does the two-axis failure coding.
\texttt{detect\_loops\_paper\_b.py} finds the longest identical action run per
cell. \texttt{grounding\_audit.py} measures identifier fidelity.
\texttt{check\_discrimination.py} reports which checks separate harnesses, and
\texttt{benchmark\_validity.py} computes task discrimination, split-half
reliability and parallel-form transfer. \texttt{rescore\_timeouts.py} produces
the rescored reading, \texttt{analyze\_two\_readings.py} builds the per-rung
quantities under both readings, and \texttt{slope\_intervals.py} puts the
task-bootstrap interval on every slope in Table~\ref{tab:span}.
\texttt{analyze\_quant\_axis.py} computes the precision ladder's level
contrasts, its slope and its Holm family.
\texttt{audit\_cells.py} gates completeness against the $18\times5\times4$ design,
and \texttt{print\_paper\_b\_numbers.py} prints every number quoted in this paper
from the summary files, so a reader can regenerate the Results section without
transcribing from a second source.

\paragraph{The analysis plan} The hypotheses, their estimators and their
refutation criteria are set out in \texttt{PREREGISTRATION\_PAPERB.md}, supplied
with the per-cell archive and frozen on 22 October 2026.

\paragraph{The study build} \system{} runs release v3.1.9 at commit
\texttt{56cd23d}. The three arms of the mechanism gradient are selected by the
\texttt{AGENT\_ABLATION} environment variable rather than by separate code
branches: \texttt{finish\_gate} disables the executable completion checks, and
\texttt{finish\_gate\_text\_only} disables those checks and the todo-sync
reminder as well. The checks switched off are the no-work refusal, the
test-not-green guard, the finish-claim artifact check, the plan-completeness
gate, the artifact-content check, and the post-finish summary. The switch is the
only difference between the arms: model, prompt, window, budgets, and the
deterministic scorer are identical, and the switch ships with the code.

\clearpage
\bibliography{refs}

\end{document}